\documentclass[letterpaper]{article} 
\usepackage[draft]{aaai2026}  
\usepackage{times}  
\usepackage{helvet}  
\usepackage{courier}  
\usepackage[hyphens]{url}  
\usepackage{graphicx} 
\usepackage{natbib}  
\usepackage{caption} 
\usepackage{algorithm}
\usepackage{algorithmic}
\usepackage{xcolor}         
\usepackage{amsmath}
\usepackage{amssymb}
\usepackage{algorithm}
\usepackage[table]{xcolor}
\usepackage{arydshln}
\usepackage{multirow}
\usepackage{booktabs}
\usepackage{subcaption}
\usepackage{newfloat}
\usepackage{listings}
\DeclareCaptionStyle{ruled}{labelfont=normalfont,labelsep=colon,strut=off} 
\floatstyle{ruled}
\newfloat{listing}{tb}{lst}{}
\floatname{listing}{Listing}
\title{SpikePool: Event-driven Spiking Transformer with Pooling Attention}
\author{
    Donghyun Lee\textsuperscript{\rm 1}, Alex Sima\textsuperscript{\rm 2}, Yuhang Li\textsuperscript{\rm 1}, Panos Stinis\textsuperscript{\rm 3}, Priyadarshini Panda\textsuperscript{\rm 1}
}
\affiliations{
    \textsuperscript{\rm 1}Department of Electrical Engineering, Yale University\\
    \textsuperscript{\rm 2}Department of Computer Science, Yale University\\
    \textsuperscript{\rm 2}Advanced Computing, Mathematics and Data Division, Pacific Northwest National Laboratory\\

    donghyun.lee@yale.edu
}

\usepackage{bibentry}

\begin{document}

\maketitle

\begin{abstract}
Building on the success of transformers, Spiking Neural Networks (SNNs) have increasingly been integrated with transformer architectures, leading to spiking transformers that demonstrate promising performance on event-based vision tasks. However, despite these empirical successes, there remains limited understanding of how spiking transformers fundamentally process event-based data. Current approaches primarily focus on architectural modifications without analyzing the underlying signal processing characteristics. In this work, we analyze spiking transformers through the frequency spectrum domain and discover that they behave as high-pass filters, contrasting with Vision Transformers (ViTs) that act as low-pass filters. This frequency domain analysis reveals why certain designs work well for event-based data, which contains valuable high-frequency information but is also sparse and noisy. Based on this observation, we propose SpikePool, which replaces spike-based self-attention with max pooling attention, a low-pass filtering operation, to create a selective band-pass filtering effect. This design preserves meaningful high-frequency content while capturing critical features and suppressing noise, achieving a better balance for event-based data processing. Our approach demonstrates competitive results on event-based datasets for both classification and object detection tasks while significantly reducing training and inference time by up to 42.5\% and 32.8\%, respectively.
\end{abstract}


\section{Introduction}
\label{sec:intro}
Event-based vision represents a fundamental departure from traditional frame-based imaging, capturing visual information as sparse, asynchronous streams of brightness changes with microsecond temporal precision rather than dense pixel arrays captured at fixed intervals~\cite{lichtsteiner2008128, posch2010qvga}. This paradigm shift from dense, synchronous frames to sparse, temporal events creates a natural synergy with Spiking Neural Networks (SNNs)~\cite{maass1997networks, roy2019towards}, which process information through discrete spike communications that mirror the temporal dynamics of biological neural systems.

The revolutionary success of transformer architectures across computer vision and natural language processing~\cite{vaswani2017attention, dosovitskiy2020image, arnab2021vivit} has naturally inspired researchers to explore their integration with SNNs, leading to the emergence of spiking transformers~\cite{zhou2022spikformer,yao2023spike,zhou2024qkformer,shi2024spikingresformer,lee2024spiking}. These architectures aim to combine the representational power of self-attention mechanisms with the energy efficiency and neuromorphic compatibility of spike-based computation. Initial results on event-based vision tasks have shown promising performance~\cite{zhang2022spiking, fang2024spiking}, suggesting that attention mechanisms can effectively capture the complex spatiotemporal dependencies inherent in event streams.

\begin{figure}[t]
\centering
\begin{tabular}{@{}c@{\hskip 0.01\linewidth}c@{\hskip 0.01\linewidth}c}
\includegraphics[width=0.6\linewidth]{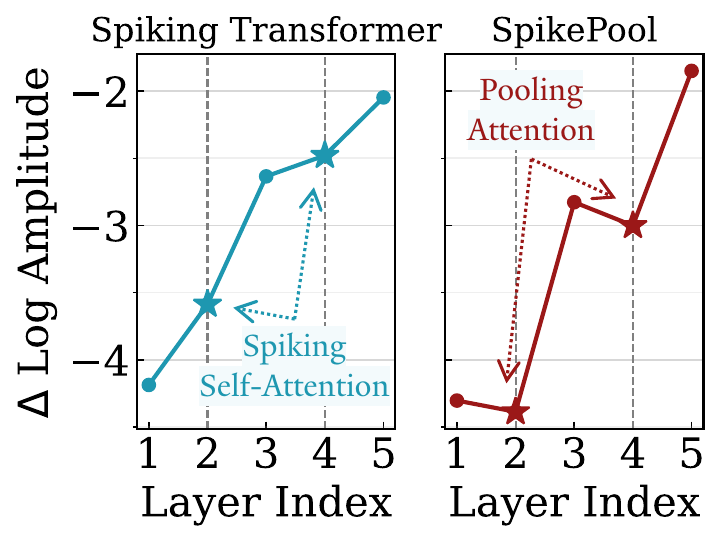} &
\includegraphics[width=0.4\linewidth]{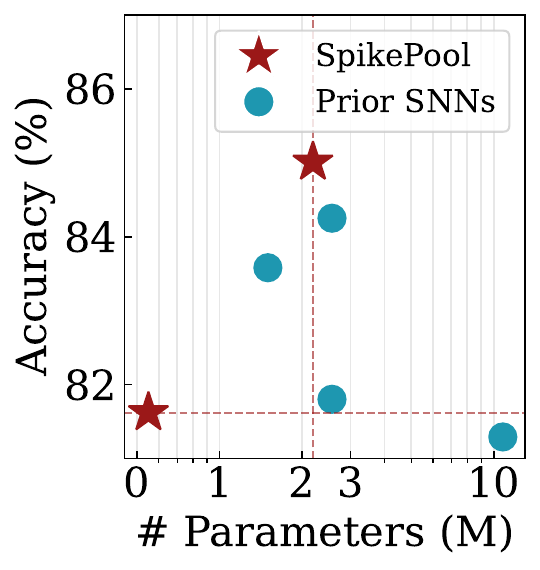}
\\
\vspace{-2mm}
{(a)} & {(b)} \\
\end{tabular}
\vspace{-1.5mm}
\caption{Frequency analysis and performance comparison of spiking transformers. (a) Relative Log Amplitude (RLA) across transformer layers, showing contrasting frequency filtering behaviors between spiking transformer and SpikePool. Layer 1 is embedding, layers 2 and 4 are attention (star-shaped), and layers 3 and 5 are Multi-Layer Perceptron (MLP) layers. (b) Accuracy and number of parameter comparison on the N-Caltech101 dataset.}
\vspace{-6mm}
\label{fig1}
\end{figure}


However, a critical gap exists in our understanding: \textit{\textbf{we lack fundamental insight into how spiking transformers process event-based data}}. Current approaches largely adopt conventional transformer mechanisms, such as Query (Q), Key (K), and Value (V) based self-attention, without analyzing whether these mechanisms are optimally suited for the unique characteristics of both spiking computation and event-based data~\cite{zhou2022spikformer,yao2023spike,lee2024spiking}. This blind adoption overlooks potential mismatches between the signal processing properties of these architectures and the nature of sparse, temporally precise event streams. Event-based data contains valuable high-frequency information~\cite{delbruck2010activity, gallego2020event} that captures rapid temporal changes, but it is also inherently sparse and noisy. Without knowledge of how spiking transformers handle such data in the frequency domain, we cannot determine whether current architectures effectively exploit these characteristics or inadvertently suppress important information while amplifying noise. To address this knowledge gap, we analyze spiking transformers through the frequency spectrum domain. Interestingly, we discover that spiking transformers behave as high-pass filters, fundamentally contrasting with Vision Transformers (ViTs) that act as low-pass filters~\cite{park2022vision, wang2022anti}. This frequency analysis motivates us to design a spiking transformer that balances frequency characteristics for optimal event-based processing.

Building on this understanding, we propose SpikePool, an event-driven spiking transformer that strategically replaces spike-based self-attention with max pooling attention. Since max pooling operates as a low-pass filter, combining it with the inherently high-pass filtering nature of spiking transformers creates a selective band-pass filtering effect as shown in Figure~\ref{fig1}(a). Rather than applying crude high-pass filtering to event data, our approach achieves a sophisticated balance through max pooling-based attention: preserving meaningful high-frequency information while capturing salient features and effectively suppressing noise. 

Our contributions are threefold: (1) We provide the first systematic frequency domain analysis of spiking transformers, revealing their high-pass filtering characteristics and explaining their behavior on event-based data. (2) We propose SpikePool, a frequency-aware architecture that strategically combines high-pass and low-pass filtering through pooling attention, creating an effective band-pass filtering mechanism for event-based vision. (3) We demonstrate that our approach achieves competitive performance for both classification and object detection tasks, while significantly improving computational efficiency (Figure~\ref{fig1}(b)), reducing training time by up to 42.5\% and inference time by up to 32.8\% compared to traditional spiking transformer.

\section{Related Works}
\label{related}
\subsection{Spiking Transformer}
The integration of transformer architectures with SNNs has emerged as a promising direction for bridging the performance gap between SNNs and ANNs while maintaining energy efficiency. Spikformer~\cite{zhou2022spikformer} pioneers this integration by adapting self-attention mechanisms for spike-based computation, primarily through binarized operations that eliminate computationally expensive Softmax normalization. Subsequent works have built upon this foundation: Spike-driven Transformer~\cite{yao2024spike} introduces masked self-attention and membrane potential-based residual connections, while Spike-driven Transformer-V2~\cite{yao2024spike2} proposes a unified framework incorporating both convolutional and self-attention operations. Recent advances include SpikingResformer~\cite{shi2024spikingresformer} and QKFormer~\cite{zhou2024qkformer}, which introduce SNN-friendly self-attention to reduce computational complexity while maintaining performance on large-scale datasets. STAtten~\cite{lee2024spiking} presents block-wise spatial-temporal attention, achieving state-of-the-art results.
Despite these empirical successes, current spiking transformers largely adopt conventional attention mechanisms from ANNs without fundamental understanding of how these architectures process information. This direct adoption overlooks potential mismatches between transformer signal processing characteristics and the unique properties of both spiking computation and event-based data, potentially leading to suboptimal designs that fail to fully exploit the advantages of neuromorphic processing.

\subsection{Frequency Analysis of Vision Transformers}
Understanding neural networks through frequency domain analysis~\cite{park2022vision,wang2022anti, bai2022improving, wang2022vtc} has provided crucial insights into their fundamental behavior.~\cite{park2022vision} demonstrates that ViT acts as a low-pass filter, preserving primarily low-frequency signals across layers, a finding that explains their robustness to high-frequency noise but sensitivity to low-frequency perturbations.~\cite{wang2022anti} further analyzes self-attention behavior in the Fourier domain, confirming the low-pass filtering characteristics and proposing methods to extract and balance high-frequency components to improve performance. These frequency domain perspectives have proven valuable for understanding why certain architectural choices succeed or fail for specific data types and have led to principled improvements in transformer design. However, no prior work has investigated the frequency domain characteristics of spiking transformers. This gap makes it difficult to design optimal spiking transformer architectures that effectively exploit the unique properties of neuromorphic data.

\begin{figure*}[t]
    \centering
    \includegraphics[width=17cm]{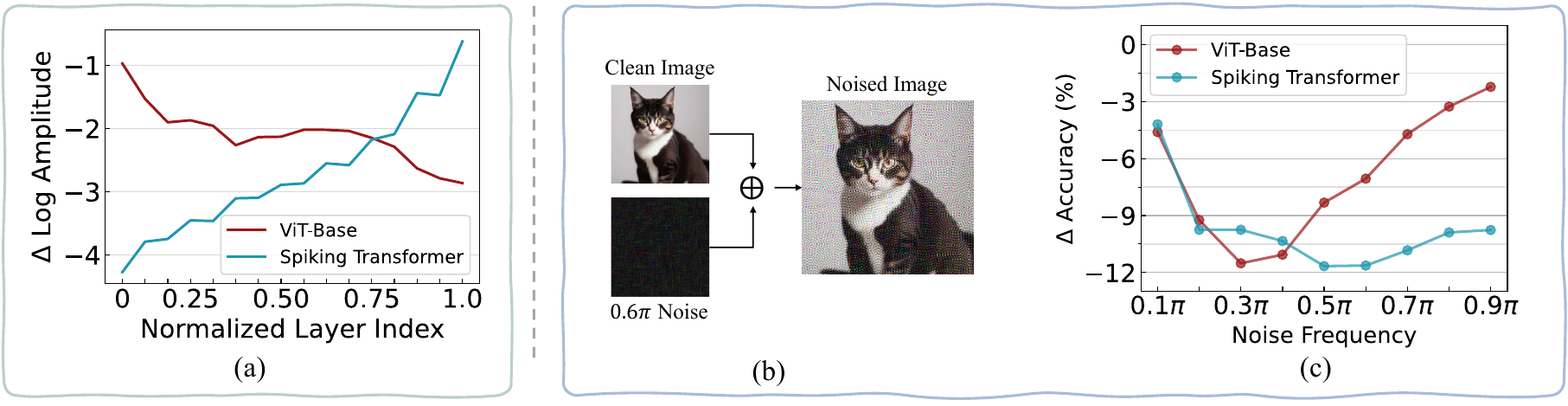}
    \vspace{-3mm}
    \caption{Frequency domain analysis of transformer architectures. (a) Relative Log Amplitude (RLA) across normalized layer depth, showing contrasting frequency filtering behaviors between ViT-Base (low-pass) and spiking transformer (high-pass). (b) Frequency-specific noise perturbation process using Fourier domain masking. (c) Robustness comparison between ViT-Base and spiking transformer.}
    \label{fig:frequency}
\vspace{-6mm}
\end{figure*}

\subsection{Event-based Vision}

Event-based vision sensors such as Dynamic Vision Sensor (DVS)~\cite{lichtsteiner2008128} and Asynchronous Time-based Image Sensor (ATIS)~\cite{posch2010qvga} generate sparse, temporally precise data that differs fundamentally from traditional dense image frames. Popular datasets including DVS128 Gesture~\cite{amir2017low}, CIFAR10-DVS~\cite{li2017cifar10}, N-MNIST~\cite{orchard2015converting}, and N-Caltech101~\cite{orchard2015converting} for classification, as well as PAFBenchmark~\cite{miao2019neuromorphic} and Gen1~\cite{cordone2022objectdetectionspikingneural} for object detection, have become standard benchmarks for neuromorphic vision. Event data exhibits natural compatibility with SNNs due to its sparse, asynchronous nature and inherent temporal dynamics~\cite{liu2018adaptive, sironi2018hats, gehrig2019end}.

For SNN processing, event data typically undergoes preprocessing through temporal binning, where continuous event streams are discretized into fixed time windows to create spike-compatible representations~\cite{liu2018adaptive, sironi2018hats, gehrig2019end}. Event-based data exhibits natural compatibility with SNNs due to several key characteristics: the sparse, asynchronous nature of event streams aligns with spike-based communication paradigms; the inherent temporal dynamics match the temporal processing capabilities of spiking neurons; and the energy-efficient generation complements low-power neuromorphic computing advantages. This synergy has made event-based datasets particularly attractive for SNN evaluation and development, leading to significant advances in neuromorphic vision applications.

\section{Preliminary}
\subsection{Leaky Integrate-and-Fire Neuron}
The Leaky Integrate-and-Fire (LIF) neuron \cite{burkitt2006review} has emerged as an important component for energy-efficient computation in SNNs~\cite{maass1997networks, roy2019towards}. The LIF neuron processes temporal information through its membrane potential dynamics as follows:
\begin{align}
    {\mathbf{u}[t+1]^l = \tau \mathbf{u}[t]^l+\mathbf{W}^lf(\mathbf{u}[t]^{l-1})}, \label{lif1} \\
    f(\mathbf{u}[t]^l) = 
    \begin{cases} 
    1 & \text{if} \; \mathbf{u}[t]^l > V_{th}, \\ 0 & \text{otherwise}
    \end{cases},
\end{align}
\noindent where, $\mathbf{u}[t]^l$ is the membrane potential in $l$-th layer at timestep $t$, $\tau\in(0,1]$ is the leaky factor for membrane potential leakage, $\mathbf{W}^l$ is the weight of $l$-th layer, and $f(\cdot)$ is the LIF function with threshold $V_{th}$. When the membrane $\mathbf{u}[t]^l$ is higher than $V_{th}$, the LIF function generates a spike and the membrane potential is reset to 0.

\subsection{Spiking Self Attention}
\label{ssa}
Standard transformers~\cite{vaswani2017attention, dosovitskiy2020image} employ self-attention mechanisms that compute attention weights through Q, K, and V operations with floating-point inputs. For input tensor $\mathbf{X}_f \in \mathbb{R}^{N \times D}$ with $N$ tokens and $D$ feature dimensions, vanilla self-attention computes $\mathbf{Q}_f = \mathbf{W_Q}\mathbf{X}_f$, $\mathbf{K}_f = \mathbf{W_K}\mathbf{X}_f$, $\mathbf{V}_f = \mathbf{W_V}\mathbf{X}_f$, followed by $\mathrm{Attn} = \texttt{Softmax}(\frac{\mathbf{Q}_f\mathbf{K}_f^\top}{\sqrt{D}})\mathbf{V}_f$ with computational complexity $\mathcal{O}(N^2D)$. Note that $\mathbf{W_Q}$, $\mathbf{W_K}$, $\mathbf{W_V} \in \mathbb{R}^{D \times D}$ are learnable linear projection matrices.

To adapt self-attention for SNNs, Spikformer~\cite{zhou2022spikformer} introduces Spiking Self Attention (SSA) with binary self-attention operations. For binary input tensor $\mathbf{X}_b \in \{0,1\}^{T \times N \times D}$ with $T$ timesteps, SSA eliminates the computationally expensive Softmax operation and formulates attention as:
\begin{equation}
\begin{aligned}
     \mathbf{Q}, \mathbf{K}, \mathbf{V} &= \texttt{LIF}(\mathbf{W_Q}\mathbf{X}_b), \texttt{LIF}(\mathbf{W_K}\mathbf{X}_b), \texttt{LIF}(\mathbf{W_V}\mathbf{X}_b), \\
     \mathrm{SSA}(\mathbf{Q},&\mathbf{K},\mathbf{V})= \texttt{LIF}(\mathbf{Q}\mathbf{K}^{\top}\mathbf{V} \cdot \alpha),
\end{aligned}
\end{equation}
where $\alpha$ is a scaling factor, and \texttt{LIF} denotes the LIF neuron function.

\section{Methodology}

\subsection{Frequency Analysis of Spiking Transformers}
We investigate spiking transformers in the Fourier domain to understand their intrinsic signal processing characteristics when handling event-based data. Following the approach of \cite{park2022vision}, we analyze the Relative Log Amplitude (RLA) of Fourier-transformed feature maps across different transformer layers.

\noindent \textbf{RLA Computation.} For each layer's feature map, we apply 2D Fourier transform and compute log amplitudes. The RLA quantifies the difference between low-frequency components (image center, 0.0$\pi$) and high-frequency components (image boundary, 1.0$\pi$) by sampling along the diagonal of the frequency spectrum. Positive RLA values indicate high-frequency emphasis (high-pass filtering), while negative values suggest preservation of low-frequency content (low-pass filtering). We compare the RLA evolution across layers between ViT-Base~\cite{dosovitskiy2020image} and Spiking Transformer (STAtten)~\cite{lee2024spiking} pre-trained on ImageNet-1K. As depicted in Figure~\ref{fig:frequency}(a), the two architectures exhibit fundamentally different frequency behaviors: while ViT-Base's RLA steadily decreases through the layers, indicating low-frequency preservation, the spiking transformer shows the opposite trend, with RLA values becoming increasingly positive, indicating high-frequency emphasis and low-frequency signal erosion.

\begin{figure}[t]
    \centering
    \includegraphics[width=\linewidth]{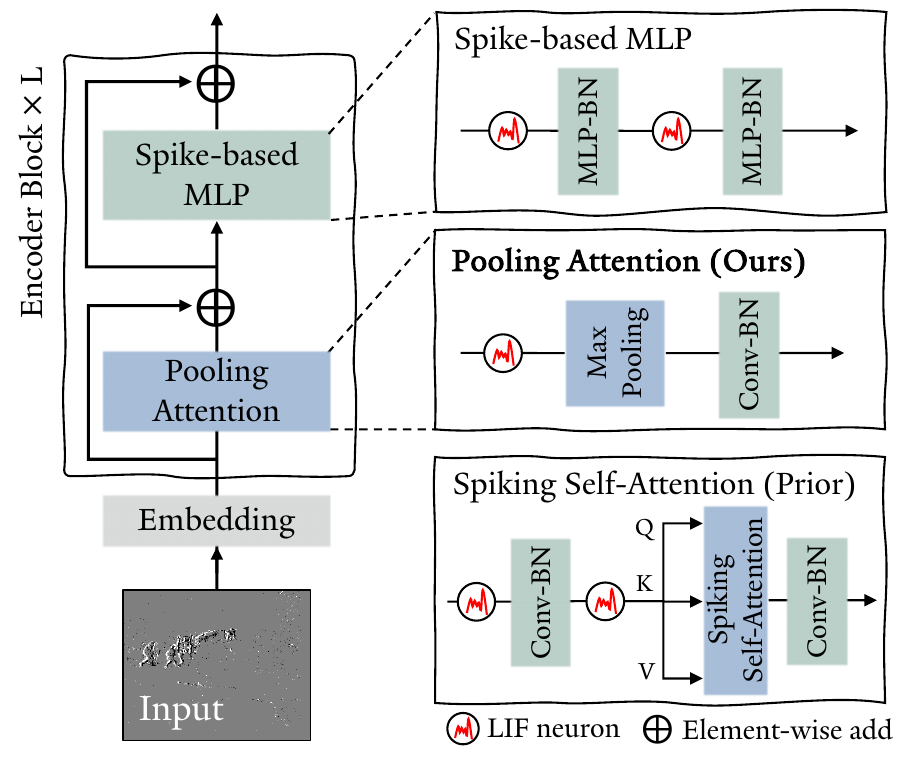}
    \caption{Overall scheme of SpikePool architecture. The pooling attention replaces prior spiking self-attention block with pooling attention. Conv, MLP, and BN represent convolution, multi-layer perception, and batch normalization.}
    \label{fig:architecture}
\vspace{-5mm}
\end{figure}

\noindent \textbf{Robustness Validation.} To validate these frequency domain findings, we conduct robustness analysis using frequency-specific noise perturbations. We generate perturbed images as $x_p = x_0 + f^{-1}(f(n) \odot M_f)$, where $x_p$ represents perturbed images, $x_0$ denotes the original image, $f$ and $f^{-1}$ are Fourier transform and inverse transform, $n$ is frequency-based random noise, and $M_f$ represents a frequency mask targeting specific frequency ranges from 0.1$\pi$ to 0.9$\pi$. The perturbation process is illustrated in Figure~\ref{fig:frequency}(b). Figure~\ref{fig:frequency}(c) confirms our frequency domain analysis through complementary robustness patterns: ViT-Base exhibits robustness to high-frequency noise but sensitivity to low-frequency perturbations, while the spiking transformer displays the opposite behavior. Specifically, when low-frequency noise (0.1$\pi$) is applied, the spiking transformer shows minimal accuracy degradation ($-$4.2\%), but performance drops significantly (10-12\%) when high-frequency noise ($>$0.5$\pi$) is introduced. The robustness results show that spiking transformers function as high-pass filters, exhibiting fundamentally different frequency characteristics compared to standard ViTs. While this high-pass behavior can preserve valuable high-frequency temporal information in event-based datasets, it may also amplify noise, motivating the need for frequency-aware architectural design.

\subsection{SpikePool}
Based on Fourier domain investigation, we propose SpikePool, which replaces typical SSA with a simple pooling attention to balance frequency processing across all encoder blocks as shown in Figure~\ref{fig:architecture}. Our baseline architecture follows~\cite{yao2023spike}, and SpikePool architecture comprises three main components: Embedding, Pooling Attention, and Spike-based Multi-Layer Perceptron (S-MLP) layers.

\subsubsection{Embedding.}
The embedding layer splits an input image into $N$ patches and extracts downsampled feature maps before attention operations. The input image $\mathbf{X} \in \mathbb{R}^{T\times C \times H \times W}$ with $T$ timesteps, $C$ input channels, $H$ height, and $W$ width, is processed through several convolution (\texttt{Conv}), batch normalization (\texttt{BN}), and max pooling (\texttt{MP}) layers as follows:
\begin{align}
    \mathbf{Y}_1&=\texttt{MP}(\texttt{LIF}(\texttt{BN}(\texttt{Conv}(\mathbf{X})))) \times 3, \\
    \mathbf{Y}_2&=\texttt{MP}(\texttt{BN}(\texttt{Conv}(\mathbf{Y}_1))), \\
    \mathbf{Y}_{out}&=\texttt{RPE}(\mathbf{Y}_2) + \mathbf{Y}_2.
\end{align}
Here, $\mathbf{Y}_{out}\in \mathbb{R}^{T\times N \times D}$ is the output of the embedding layer and \texttt{RPE} represents Relative Position Embedding (RPE), which consists of one \texttt{Conv} and \texttt{BN} layer. The output $\mathbf{Y}_{out}$ can also be considered as the membrane potential, so spike outputs are fired in the following encoder block.

\begin{figure}[t]
    \centering
    \includegraphics[width=0.9\linewidth]{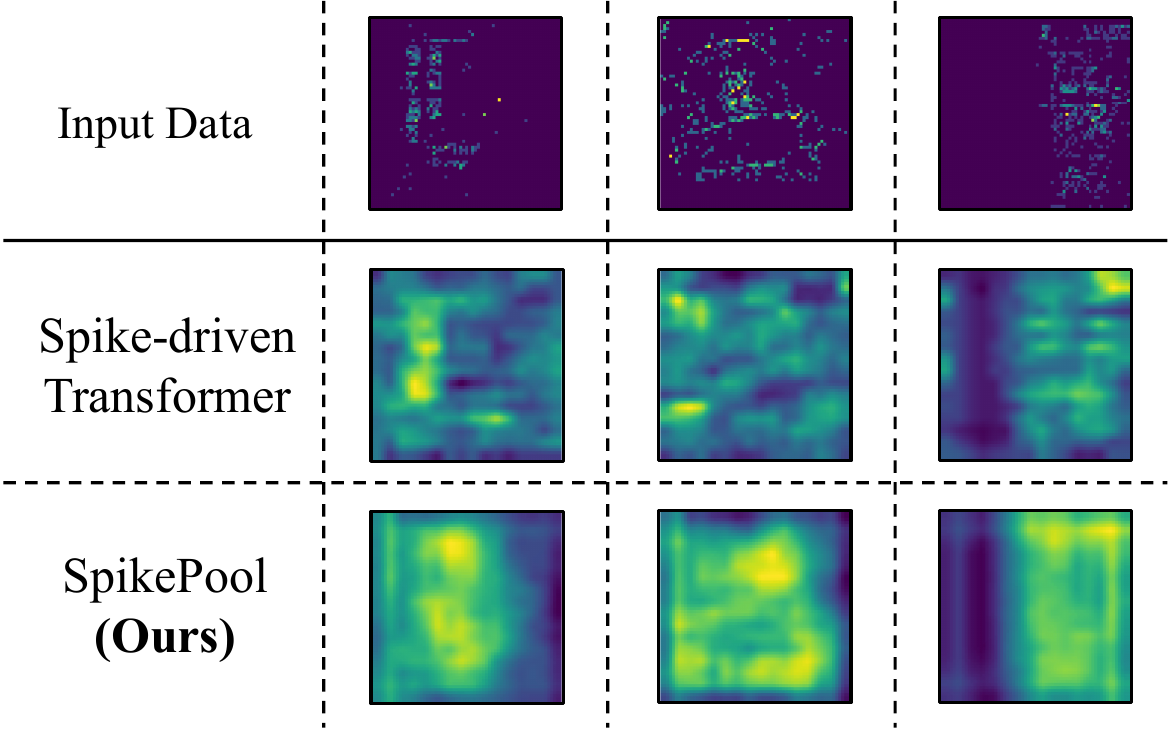}
    \vspace{-3mm}
    \caption{Attention feature maps comparison. Input event data (top), spike-driven transformer~\cite{yao2023spike} output (middle), and SpikePool output (bottom). }
    \label{fig:feature}
\vspace{-4mm}
\end{figure}

\begin{table*}[t]
\small
  \caption{Performance comparison between our methods and previous works on CIFAR10-DVS and N-Caltech101 datasets. In the architecture column, $L$-$D$ represents $L$ number of encoder blocks and $D$ hidden dimensions.}
  \vspace{-4pt}
  \label{dvs_result}
  \centering
  \begin{tabular}{llcccc}
    \toprule
    Method & Architecture & \#Param (M) & Timestep & CIFAR10-DVS ($\%$) & N-Caltech101 ($\%$)      \\
    \midrule
     tdBN \cite{zheng2021going} & ResNet19 &12.54 & 10 & 67.80 & - \\
     PLIF \cite{fang2021incorporating} & ConvNet & 36.71 & 20 & 74.80 & - \\
     Dspike \cite{li2021differentiable} & ResNet18&11.21 & 10 & 75.40 & - \\
     DSR \cite{meng2022training} & ResNet18 & 11.21& 20 & 77.27 & - \\
     TT-SNN \cite{lee2024tt} & ResNet34 & 20.82 & 6 & - & 77.80 \\
     NDA \cite{li2022neuromorphic} & VGG11 & 129.17 &10 & 79.6 & 78.2 \\
     \cdashline{1-6}\\[-1.75ex]
     Spikformer \cite{zhou2022spikformer} & 2-256 & 2.57& 16 & 80.9 & - \\
     Spike-driven Transformer \cite{yao2024spike}  & 2-256&2.57 & 16 & 80.0 & 81.80 \\
     SpikingResformer \cite{shi2024spikingresformer}   & 4-384 & 10.79& 16 & 78.80 & 81.29  \\
     QKFormer \cite{zhou2024qkformer}  & 2-256 & 1.50& 16 & 82.90 & 83.58  \\
     STAtten \cite{lee2024spiking}  & 2-256 & 2.57 &16 & 83.90 & 84.25 \\
     \cdashline{1-6}\\[-1.75ex]
     \cellcolor{gray!25}\textbf{SpikePool-S} &  \cellcolor{gray!25}2-128 & \cellcolor{gray!25}\textbf{0.55} &\cellcolor{gray!25}16& \cellcolor{gray!25}\textbf{80.20}& \cellcolor{gray!25}\textbf{81.62} \\
     \cellcolor{gray!25}\textbf{SpikePool-B} &  \cellcolor{gray!25}2-256 & \cellcolor{gray!25} \textbf{2.19} &\cellcolor{gray!25}16& \cellcolor{gray!25}\textbf{82.70}& \cellcolor{gray!25}\textbf{85.01} \\
     
    \bottomrule
    \vspace{-6mm}
  \end{tabular}
\end{table*}

\subsubsection{Pooling Attention.}
Instead of the computationally intensive SSA mechanism, we propose a simplified pooling attention that maintains effective feature aggregation while reducing computational overhead. As illustrated in Figure~\ref{fig:architecture}, our pooling attention consists of a 2D max pooling operation followed by convolutional and batch normalization layers. For the input feature tensor $\mathbf{Y}_{out} \in \mathbb{R}^{T \times N \times D}$ from the embedding layer, the pooling attention operates as follows:
\begin{align}
    \mathbf{Z}_{lif} &= \texttt{LIF}(\mathbf{Y}_{out}), \\
    \mathbf{Z}_{pool} &= \texttt{MP}(\mathbf{Z}_{lif}), \\
    \mathbf{Z}_{out} &= \texttt{Conv-BN}(\mathbf{Z}_{pool}) + \mathbf{Y}_{out},
\end{align}
where $\texttt{MP}$ performs 2D max pooling across spatial dimensions to capture the most salient features, $\texttt{Conv-BN}$ represents the convolution with batch normalization layers defined as $\texttt{BN}(\texttt{Conv}(\cdot))$, and the residual connection preserves the original input information. This design eliminates the quadratic complexity of traditional self-attention while preserving the ability to aggregate global information across spatial locations. Crucially, the max pooling operation provides complementary frequency characteristics to the inherent high-pass filtering of spiking neurons, creating a more balanced frequency response that both captures salient features and suppresses noise, as demonstrated in our frequency analysis (Figure~\ref{fig1}(a)).

\subsubsection{Spike-based MLP.}
The S-MLP follows the pooling attention module and provides nonlinear transformation capabilities. The S-MLP consists of two consecutive MLP layers with batch normalization, connected through a residual pathway. For the input feature tensor $\mathbf{Z}_{out} \in \mathbb{R}^{T \times N \times D}$ from the pooling attention, the S-MLP operates as follows:
\begin{align}
    \mathbf{H}_1 &= \texttt{MLP-BN}(\texttt{LIF}(\mathbf{Z}_{out})), \\
    \mathbf{H}_2 &= \texttt{MLP-BN}(\texttt{LIF}(\mathbf{H}_1)), \\
    \mathbf{H}_{out} &= \mathbf{H}_2 + \mathbf{Z}_{out},
\end{align}
where $\texttt{MLP-BN}$ represents multi-layer perceptron with a batch normalzation layer defined as $\texttt{BN}(\texttt{MLP}(\cdot))$. Similar to~\cite{yao2023spike}, we implement the MLP using convolution operations. Following this module, a classification head computes the class probabilities for the final classification.

\subsection{Feature Map Analysis}
To qualitatively verify the effectiveness of the proposed SpikePool, we visualize and analyze the feature maps to understand how different spiking transformers process input data. Figure~\ref{fig:feature} shows the attention features from both the spike-driven transformer~\cite{yao2023spike} and SpikePool architecture.

Consistent with our frequency domain investigation, SpikePool effectively extracts the low-frequency components (main features) with a zoom-in effect while filtering out high-frequency noise pixels. In contrast, the spike-driven transformer captures both main features and background noise, leading to degraded representation quality. This difference is particularly evident in the second column, where the spike-driven transformer fails to capture the main object effectively, while SpikePool maintains clear focus on the primary features. This visualization confirms that our pooling attention mechanism successfully acts as a low-pass filter and demonstrates better feature selectivity than traditional SSA.

\section{Experiments}
\label{experiments}

\begin{table*}[t]
\small
  \centering
  \setlength{\tabcolsep}{4pt}
  \caption{Comparison between SpikePool and SpikingViT-B on PAFBenchmark and Gen1. $\dagger$ represents our implementation for fair comparison. "D" in the timestep column represents the dynamic timestep, which is discussed in Appendix B.}
  \vspace{-5pt}
  \label{obj_full}
  \begin{tabular}{@{}l c c c ccc cc cc@{}}
    \toprule
    \multicolumn{1}{c}{Method} & \multicolumn{1}{c}{Type} & \multicolumn{1}{c}{Model} & \multicolumn{1}{c}{Timestep} & \multicolumn{1}{c}{Params(M)} 
      & \multicolumn{2}{c}{PAF}
      
      & \multicolumn{2}{c}{Gen1} & \multicolumn{2}{c}{1Mpx} \\
    \cmidrule(lr){6-7} \cmidrule(lr){8-9} \cmidrule(lr){10-11}& & & &
     & mAP & AP\textsubscript{50}
     & mAP & AP\textsubscript{50}
     & mAP & AP\textsubscript{50}


     \\
    \midrule
    MatrixLSTM    & ANN & RNN + CNN + YOLOv3 & - &  61.5    & – & – & 0.310 & – &  - & -  \\
    YOLOv3 Events & ANN & CNN + YOLOv3 & - &\textgreater 60 & – & – & 0.312 & –  &  0.346 & -    \\
    RED   & ANN &RNN + CNN + SSD~\cite{liu2016ssd}& - & 24.1    & – &  – & 0.400 & – & 0.430 & -     \\
    ASTMNet     & ANN  &RNN + TCNN + SSD & - & \textgreater 100    & – & – & 0.467 & – &  \textbf{0.483} & -    \\
    RVT          & ANN &Transformer + RNN + YOLOX& - &18.5  & – & – & \textbf{0.472} & \textbf{0.701} &  0.474 & \textbf{0.723}  \\
    \midrule
    LT-SNN& SNN & YOLOv2 & - & –  & – & – & 0.298 & – &  - & -    \\
    KD-SNN& SNN & CenterNet & 5 & 12.97 & – & – & 0.229 & – &  - & -   \\
    EMS-ResNet34 & SNN & EMS-Res + YOLOv3  & 5 & 14.4  & – & – & 0.310 & 0.590 &  - & -  \\
    EAS-SNN      & SNN  & YOLOX & 3 & 25.3 & – & – & \textbf{0.409} & \textbf{0.731} &  - & -    \\
    SpikingViT-B  & SNN  &Spikformer + YOLOX& D & 21.5  & 0.619$\dagger$ & 0.921$\dagger$ & 0.379$\dagger$ & 0.609$\dagger$  &  0.330$\dagger$ & 0.562$\dagger$    \\
    \cdashline{1-11}\\[-1.75ex]
     
    \cellcolor{gray!25}\textbf{SpikePool}   & \cellcolor{gray!25}SNN    & \cellcolor{gray!25}Spikformer + YOLOX  & \cellcolor{gray!25}D &\cellcolor{gray!25}\textbf{19.3} & \cellcolor{gray!25}\textbf{0.641} & \cellcolor{gray!25}\textbf{0.926} & \cellcolor{gray!25}\textbf{0.393} & \cellcolor{gray!25}\textbf{0.632} & \cellcolor{gray!25}\textbf{0.347} & \cellcolor{gray!25}\textbf{0.605}  \\
    \bottomrule
  \end{tabular}
  \vspace{-8pt}
\end{table*}

\subsection{Classification}
We evaluate SpikePool on two neuromorphic datasets: CIFAR10-DVS~\cite{li2017cifar10} and N-Caltech101~\cite{orchard2015converting}. CIFAR10-DVS contains 10,000 samples with 9,000 training and 1,000 testing samples, captured by a DVS camera from static CIFAR10 images. N-Caltech101 consists of 8,831 DVS data samples, also captured by DVS camera from the Caltech101 dataset~\cite{li2022caltech}. Following the experimental setup of~\cite{li2022neuromorphic, lee2024spiking}, we resize the data to 64$\times$64 and apply data augmentation strategies including horizontal flip, rolling, rotation, and CutMix~\cite{yun2019cutmix}. All experiments are trained from scratch, and detailed hyperparameters are provided in Appendix A.

For the classification task, we provide two different architecture sizes: SpikePool-S and SpikePool-B. Both variants have the same number of layers but differ in hidden dimension size, where SpikePool-S uses smaller dimensions (128) and SpikePool-B uses larger dimensions (256). The experimental results are shown in Table~\ref{dvs_result}. Our proposed SpikePool achieves competitive performance across both datasets. SpikePool-B attains 82.70\% accuracy on CIFAR10-DVS and 85.01\% accuracy on N-Caltech101, outperforming most existing methods while maintaining significantly lower parameter counts. Notably, SpikePool-B achieves state-of-the-art performance on N-Caltech101 among architectures trained from scratch. Furthermore, SpikePool-S uses only 0.55M parameters while achieving competitive performance (80.20\% on CIFAR10-DVS and 81.62\% on N-Caltech101). This significant parameter reduction validates our approach of replacing computationally expensive self-attention with efficient pooling operations, achieving an optimal trade-off between model performance and computational efficiency.

\subsection{Object Detection}

For object detection, we evaluate SpikePool on three event datsets: PAFBenchmark (PAF)~\cite{miao2019neuromorphic}, Gen1~\cite{de2020large}, and 1Mpx~\cite{perot2020learningdetectobjects1}. PAF consists of event‐based pedestrian detection sequences recorded by a DAVIS346 camera, containing 4,670 frame images with varying densities of pedestrians under different lighting/weather conditions. Gen1 is a large‐scale event‐based dataset captured by a 304$\times$240 ATIS sensor~\cite{de2020large}, featuring a total of 228,000 car and 28,000 pedestrian bounding boxes for object detection in diverse driving scenarios. 1Mpx is a high-resolution, large-scale event-based dataset captured on a 1 megapixel event camera ~\cite{perot2020learningdetectobjects1} with 25M bounding boxes of cars, pedestrians, and two-wheelers, labeled at high frequency. Following the experimental setup of ~\cite{yu2024spikingvit},  bounding boxes with side lengths less than 20 pixels (10 pixels for Gen1) or a diagonal length less than 60 pixels (30 pixels for Gen1) have been excluded. We adopt the overall detection framework of SpikingViT~\cite{yu2024spikingvit}, which integrates spikformer~\cite{zhou2022spikformer} encoder composed of SSA blocks with a YOLOX-style~\cite{ge2021yolox} detection head. A brief introduction to SpikingViT architecture is provided in Appendix B. To create SpikePool, we retain the detection head from SpikingViT-B, but replace each SSA block in the encoder with a pooling attention block. We train both SpikePool and SpikingViT from scratch, and the experimental results are shown in Table~\ref{obj_full}. The previous works for comparison are as follows: MatrixLSTM~\cite{cannici2019asynchronous}, YOLOv3 Events~\cite{jiang2019mixed}, RED~\cite{perot2020learning}, ASTMNet~\cite{li2022asynchronous}, RVT~\cite{gehrig2023recurrent}, 
LT-SNN~\cite{hasssan2023lt}, KD-SNN~\cite{bodden2024spiking}, EMS-ResNet34~\cite{su2023deep}, EAS-SNN~\cite{wang2024eas}, and SpikingViT~\cite{yu2024spikingvit}. Detailed hyperparameters are shown in Appendix A. 
\begin{figure}[t]
    \centering
    \includegraphics[width=\linewidth]{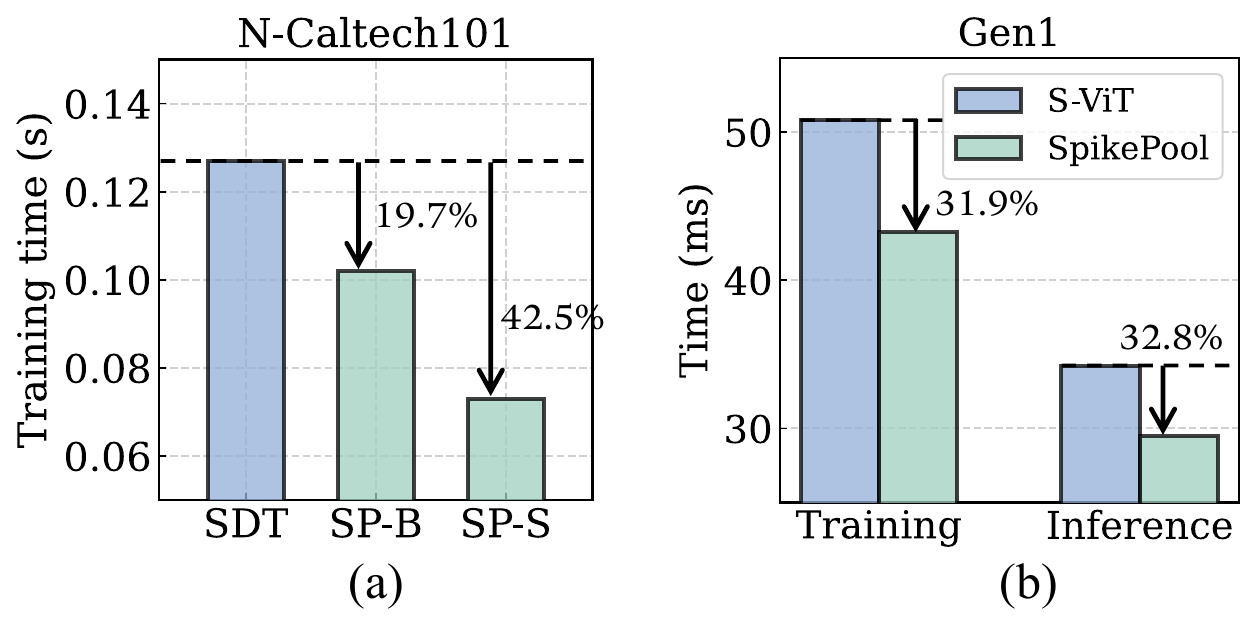}
    \vspace{-8mm}
    \caption{Training and inference time comparison on N-Caltech101 and Gen1 datasets. (a) N-Caltech101 showing training time per iteration for Spike-driven Transformer (SDT), SpikePool-B (SP-B), and SpikePool-S (SP-S). (b) Gen1 dataset comparing training and inference times for SpikingViT (S-ViT) and SpikePool (SP).}
    \label{fig:runtime}
\vspace{-6mm}
\end{figure}

\begin{figure*}[t]
  \centering
  \setlength{\tabcolsep}{10pt} 
  \begin{tabular}{@{}c@{\hspace{4pt}}c@{\hspace{4pt}}c@{\hspace{4pt}}c@{\hspace{4pt}}c@{}}
    \begin{minipage}[b]{.03\linewidth}
      \rotatebox{90}{\footnotesize SpikingViT}\\[60pt]
      \rotatebox{90}{\footnotesize SpikePool}\\[55pt]
      \rotatebox{90}{\footnotesize Ground Truth}\\[25pt]
    \end{minipage}
    \begin{subfigure}[b]{0.23\linewidth}
      \includegraphics[width=\linewidth]{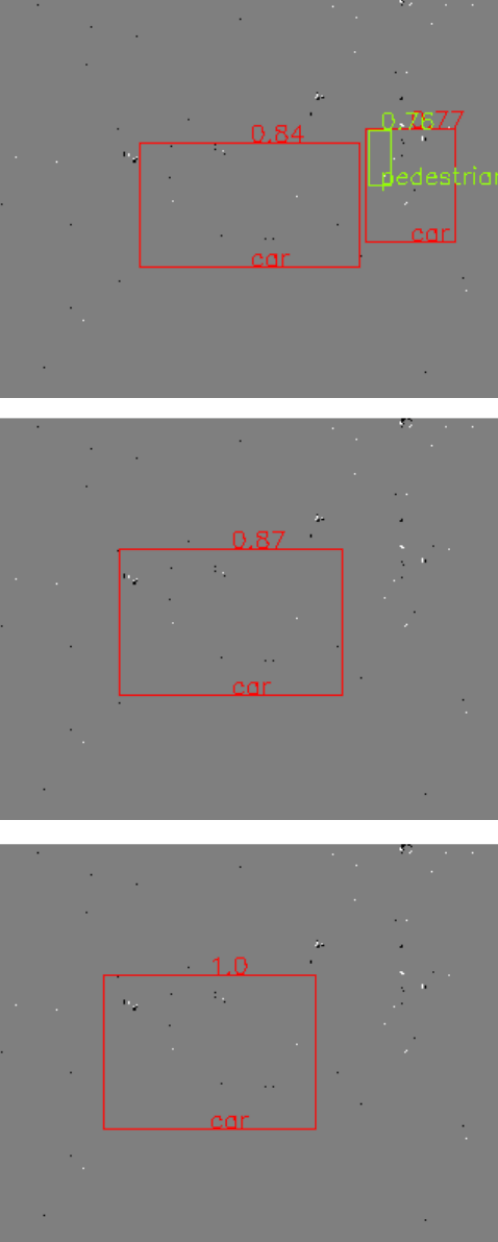}
      \caption{Example 1}
      \label{fig:spikepool_gen1_1}
    \end{subfigure} &
    \begin{subfigure}[b]{0.23\linewidth}
      \includegraphics[width=\linewidth]{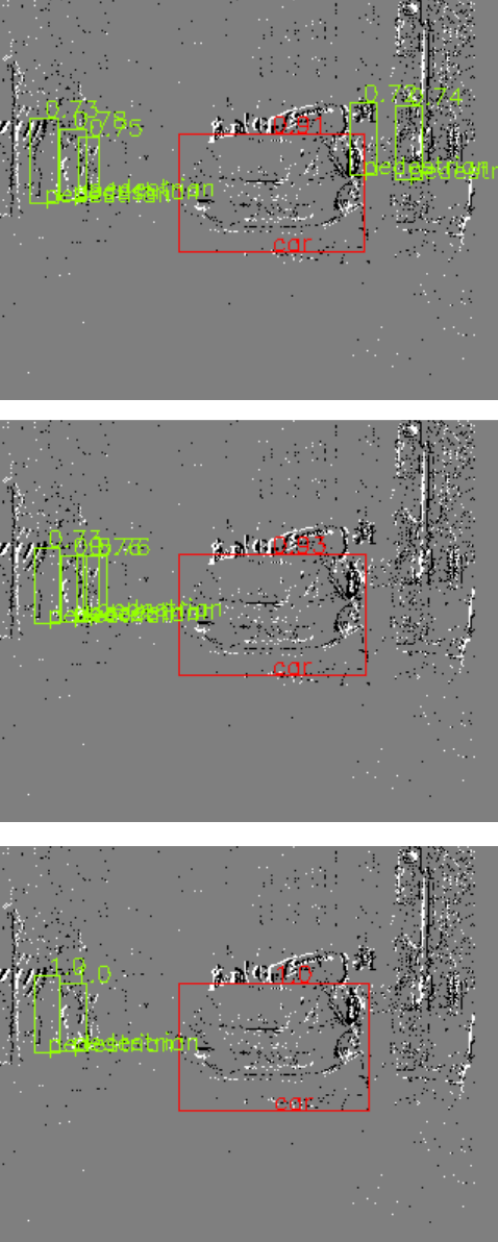}
      \caption{Example 2}
      \label{fig:spikepool_gen1_2}
    \end{subfigure} &
    \begin{subfigure}[b]{0.23\linewidth}
      \includegraphics[width=\linewidth]{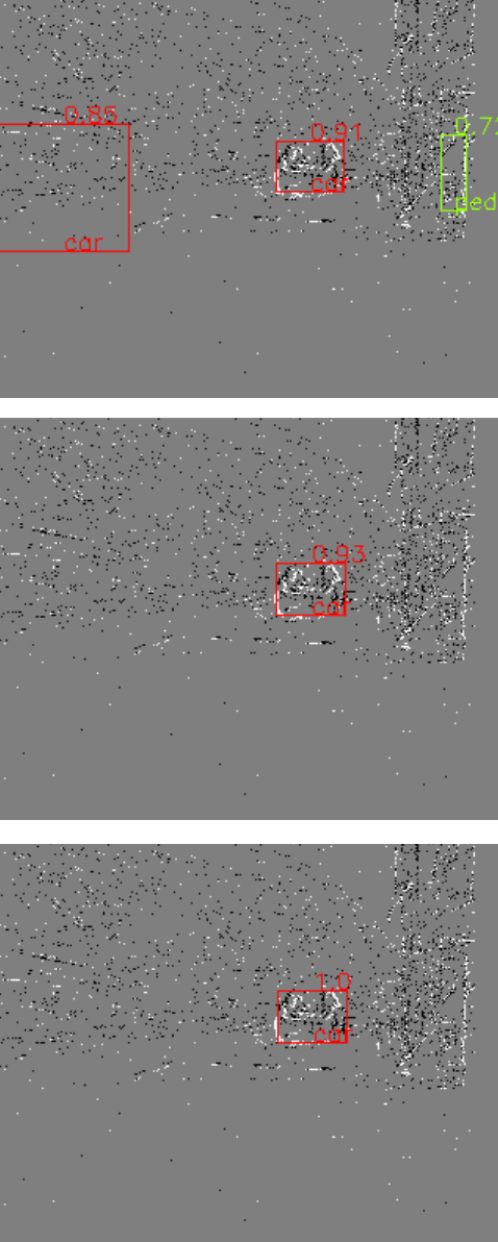}
      \caption{Example 3}
      \label{fig:spikepool_gen1_3}
    \end{subfigure} &
    \begin{subfigure}[b]{0.23\linewidth}
      \includegraphics[width=\linewidth]{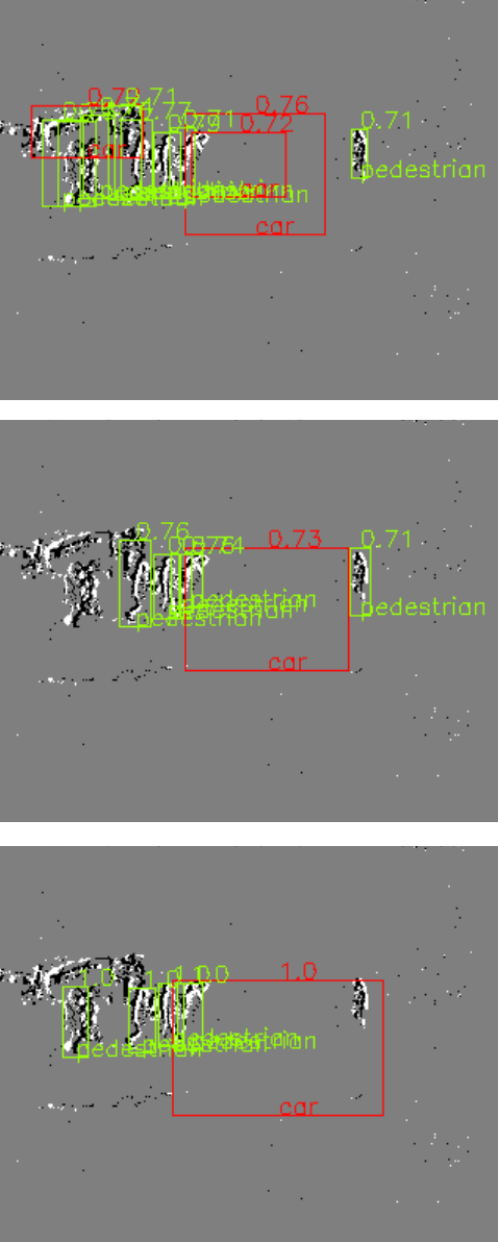}
      \caption{Example 4}
      \label{fig:spikepool_gen1_4}
    \end{subfigure}
  \end{tabular}
\vspace{-5pt}
  \caption{SpikePool, SpikingViT, and Ground Truth visualizations on the Gen1 dataset. The left‐hand labels indicate whether the top, middle, and bottom portions of each image are network outputs and ground truth, respectively. For failure case examples, see Appendix E.}
  \label{fig:spikepool_gen1_viz}
  \vspace{-5pt}
\end{figure*}

SpikePool outperforms SpikingViT-B across multiple datasets, achieving 0.641 mAP and 0.926 AP\textsubscript{50} on PAFBenchmark, 0.393 mAP and 0.632 AP\textsubscript{50} on Gen1, and 0.347 mAP and 0.605 AP\textsubscript{50} on 1Mpx, representing the comparable performance among SNN-based approaches. In addition, SpikePool achieves competitive performance with fewer parameters than SpikingViT-B, using 19.3M parameters compared to SpikingViT's 21.5M, a decrease of 10\%. This aligns with our classification experiment findings, where replacing computationally expensive self-attention mechanics with pooling operations leads to better model efficiency. Figure~\ref{fig:spikepool_gen1_viz} shows the visualizations of bounding-box predictions from both SpikingViT and SpikePool along with their ground truth labels.

\subsection{Training and Inference Time}

To demonstrate computational efficiency, we measure training and inference times across different tasks compared to baseline spiking transformer approaches. Figure~\ref{fig:runtime} shows runtime performance on N-Caltech101 for classification and Gen1 for object detection using A5000 and A100 GPUs. 

On N-Caltech101, our SpikePool variants achieve significant speedup in training time compared to the Spike-driven Transformer (SDT)~\cite{yao2023spike}. SpikePool-B reduces training time from 0.127s to 0.102s per batch update (19.7\%~$\downarrow$), while SpikePool-S further accelerates training to 0.073s (42.5\%~$\downarrow$). On the Gen1 dataset, we compare both training and inference speed of the encoder part per iteration against SpikingViT (S-ViT). SpikePool achieves substantial improvements in both phases: training time is reduced from 50.84ms to 34.19ms (31.9\%~$\downarrow$), and inference time is accelerated from 43.28ms to 29.49ms (32.8\%~$\downarrow$), showing consistent speedup across different datasets. 




\begin{table}[t]
\footnotesize
  \centering
\caption{Performance improvement when integrating pooling attention into existing spiking transformers.}
  \vspace{-7pt}
  \label{tab:plug}
  \begin{tabular}{lcc}
    \toprule
    Method & CIFAR10-DVS & N-Caltech101 \\
    \midrule
    Spikformer & 80.9 \% & 82.71 \%\\
    \cellcolor{gray!25}+ Pooling Attention & \cellcolor{gray!25}\textbf{81.2} \% & \cellcolor{gray!25}\textbf{83.58} \%\\
    \cdashline{1-3}\\[-1.75ex]
    QKFormer & 82.9 \% & 83.58 \% \\
    \cellcolor{gray!25}+ Pooling Attention & \cellcolor{gray!25}\textbf{84.7} \% & \cellcolor{gray!25}\textbf{84.79} \% \\
    \bottomrule
  \end{tabular}
  \vspace{-8pt}
\end{table}

\subsection{Generalization}
While our SpikePool architecture is based on the Spike-driven Transformer~\cite{yao2023spike}, we investigate the broader applicability of our pooling attention by integrating it into other existing spiking transformer architectures. To demonstrate generalization capability, we plug our pooling attention into Spikformer~\cite{zhou2022spikformer} and QKFormer~\cite{zhou2024qkformer}, replacing their original self-attention modules.

As shown in Table~\ref{tab:plug}, our pooling attention consistently improves performance across different base architectures. These consistent improvements across diverse spiking transformer architectures validate that max pooling is effectively aligned with the fundamental characteristics of spike-based processing, regardless of the specific transformer design. Implementation details are provided in Appendix C.

\begin{table}[t]
\footnotesize
  \centering
  \caption{Accuracy comparison with different pooling strategies.}
  \vspace{-7pt}
  \label{tab:pooling_ablation}
  \begin{tabular}{lcc}
    \toprule
    Method & CIFAR10-DVS (\%) & N-Caltech101 (\%) \\
    \midrule
    2D Avg Pooling & 79.6 & 81.51 \\
    3D Max Pooling & 80.2 & 83.04\\
    \cellcolor{gray!25}2D Max Pooling & \cellcolor{gray!25}\textbf{82.7} & \cellcolor{gray!25}\textbf{85.01} \\
    \bottomrule
  \end{tabular}
  \vspace{-16pt}
\end{table}

\vspace{-2pt}
\subsection{Different Pooling Operations}
\vspace{-1pt}
SpikePool employs 2D max pooling, which performs pooling operations exclusively in the spatial domain. To validate the effectiveness of our pooling choice, we conduct experiments with alternative pooling strategies: 2D average pooling and 3D max pooling. 2D average pooling aggregates values by computing the mean within each kernel region, while 3D max pooling extends the max pooling operation to include the temporal dimension. As shown in Table~\ref{tab:pooling_ablation}, 2D max pooling achieves the highest performance among all variants. We attribute this high accuracy to two key factors. First, average pooling tends to aggregate nearby noise information rather than selecting the most salient features. Second, 3D max pooling disrupts temporal dynamics by over-selecting sporadic spikes across time steps.


\vspace{-3pt}
\section{Conclusion}
We present SpikePool, a simple and efficient spiking transformer architecture designed specifically for event-based vision tasks. Through comprehensive frequency domain analysis, we reveal that spiking transformers function as high-pass filters, differing from standard Vision Transformers that act as low-pass filters. This insight motivates our pooling-based attention mechanism, which replaces computationally expensive self-attention with efficient max pooling operations while achieving better frequency domain balance. Our approach demonstrates significant improvements across multiple event-based datasets, achieving competitive accuracy with reduced computational overhead. The frequency analysis framework provides principled insights for future neuromorphic architecture design, while the SpikePool architecture offers a practical solution for efficient event-based processing.

\clearpage
\bibliography{aaai2026}

\begin{thebibliography}{54}
\providecommand{\natexlab}[1]{#1}

\bibitem[{Amir et~al.(2017)Amir, Taba, Berg, Melano, McKinstry, Di~Nolfo, Nayak, Andreopoulos, Garreau, Mendoza et~al.}]{amir2017low}
Amir, A.; Taba, B.; Berg, D.; Melano, T.; McKinstry, J.; Di~Nolfo, C.; Nayak, T.; Andreopoulos, A.; Garreau, G.; Mendoza, M.; et~al. 2017.
\newblock A low power, fully event-based gesture recognition system.
\newblock In \emph{Proceedings of the IEEE conference on computer vision and pattern recognition}, 7243--7252.

\bibitem[{Arnab et~al.(2021)Arnab, Dehghani, Heigold, Sun, Lu{\v{c}}i{\'c}, and Schmid}]{arnab2021vivit}
Arnab, A.; Dehghani, M.; Heigold, G.; Sun, C.; Lu{\v{c}}i{\'c}, M.; and Schmid, C. 2021.
\newblock Vivit: A video vision transformer.
\newblock In \emph{Proceedings of the IEEE/CVF international conference on computer vision}, 6836--6846.

\bibitem[{Bai et~al.(2022)Bai, Yuan, Xia, Yan, Li, and Liu}]{bai2022improving}
Bai, J.; Yuan, L.; Xia, S.-T.; Yan, S.; Li, Z.; and Liu, W. 2022.
\newblock Improving vision transformers by revisiting high-frequency components.
\newblock In \emph{European conference on computer vision}, 1--18. Springer.

\bibitem[{Bodden et~al.(2024)Bodden, Ha, Schwaiger, Kreuzberg, and Behnke}]{bodden2024spiking}
Bodden, L.; Ha, D.~B.; Schwaiger, F.; Kreuzberg, L.; and Behnke, S. 2024.
\newblock Spiking centernet: A distillation-boosted spiking neural network for object detection.
\newblock In \emph{2024 International Joint Conference on Neural Networks (IJCNN)}, 1--9. IEEE.

\bibitem[{Burkitt(2006)}]{burkitt2006review}
Burkitt, A.~N. 2006.
\newblock A review of the integrate-and-fire neuron model: I. Homogeneous synaptic input.
\newblock \emph{Biological cybernetics}, 95: 1--19.

\bibitem[{Cannici et~al.(2019)Cannici, Ciccone, Romanoni, and Matteucci}]{cannici2019asynchronous}
Cannici, M.; Ciccone, M.; Romanoni, A.; and Matteucci, M. 2019.
\newblock Asynchronous convolutional networks for object detection in neuromorphic cameras.
\newblock In \emph{Proceedings of the IEEE/CVF Conference on Computer Vision and Pattern Recognition Workshops}, 0--0.

\bibitem[{Cordone, Miramond, and Thierion(2022)}]{cordone2022objectdetectionspikingneural}
Cordone, L.; Miramond, B.; and Thierion, P. 2022.
\newblock Object Detection with Spiking Neural Networks on Automotive Event Data.
\newblock \emph{arXiv preprint arXiv:2205.04339}.

\bibitem[{De~Tournemire et~al.(2020)De~Tournemire, Nitti, Perot, Migliore, and Sironi}]{de2020large}
De~Tournemire, P.; Nitti, D.; Perot, E.; Migliore, D.; and Sironi, A. 2020.
\newblock A large scale event-based detection dataset for automotive.
\newblock \emph{arXiv preprint arXiv:2001.08499}.

\bibitem[{Delbr{\"u}ck et~al.(2010)Delbr{\"u}ck, Linares-Barranco, Culurciello, and Posch}]{delbruck2010activity}
Delbr{\"u}ck, T.; Linares-Barranco, B.; Culurciello, E.; and Posch, C. 2010.
\newblock Activity-driven, event-based vision sensors.
\newblock In \emph{Proceedings of 2010 IEEE international symposium on circuits and systems}, 2426--2429. IEEE.

\bibitem[{Dosovitskiy et~al.(2020)Dosovitskiy, Beyer, Kolesnikov, Weissenborn, Zhai, Unterthiner, Dehghani, Minderer, Heigold, Gelly et~al.}]{dosovitskiy2020image}
Dosovitskiy, A.; Beyer, L.; Kolesnikov, A.; Weissenborn, D.; Zhai, X.; Unterthiner, T.; Dehghani, M.; Minderer, M.; Heigold, G.; Gelly, S.; et~al. 2020.
\newblock An image is worth 16x16 words: Transformers for image recognition at scale.
\newblock \emph{arXiv preprint arXiv:2010.11929}.

\bibitem[{Fang et~al.(2023)Fang, Chen, Ding, Yu, Masquelier, Chen, Huang, Zhou, Li, and Tian}]{fang2023spikingjelly}
Fang, W.; Chen, Y.; Ding, J.; Yu, Z.; Masquelier, T.; Chen, D.; Huang, L.; Zhou, H.; Li, G.; and Tian, Y. 2023.
\newblock Spikingjelly: An open-source machine learning infrastructure platform for spike-based intelligence.
\newblock \emph{Science Advances}, 9(40): eadi1480.

\bibitem[{Fang et~al.(2021)Fang, Yu, Chen, Masquelier, Huang, and Tian}]{fang2021incorporating}
Fang, W.; Yu, Z.; Chen, Y.; Masquelier, T.; Huang, T.; and Tian, Y. 2021.
\newblock Incorporating learnable membrane time constant to enhance learning of spiking neural networks.
\newblock In \emph{Proceedings of the IEEE/CVF international conference on computer vision}, 2661--2671.

\bibitem[{Fang et~al.(2024)Fang, Wang, Zhang, Cao, Chen, and Xu}]{fang2024spiking}
Fang, Y.; Wang, Z.; Zhang, L.; Cao, J.; Chen, H.; and Xu, R. 2024.
\newblock Spiking wavelet transformer.
\newblock In \emph{European Conference on Computer Vision}, 19--37. Springer.

\bibitem[{Gallego et~al.(2020)Gallego, Delbr{\"u}ck, Orchard, Bartolozzi, Taba, Censi, Leutenegger, Davison, Conradt, Daniilidis et~al.}]{gallego2020event}
Gallego, G.; Delbr{\"u}ck, T.; Orchard, G.; Bartolozzi, C.; Taba, B.; Censi, A.; Leutenegger, S.; Davison, A.~J.; Conradt, J.; Daniilidis, K.; et~al. 2020.
\newblock Event-based vision: A survey.
\newblock \emph{IEEE transactions on pattern analysis and machine intelligence}, 44(1): 154--180.

\bibitem[{Ge et~al.(2021)Ge, Liu, Wang, Li, and Sun}]{ge2021yolox}
Ge, Z.; Liu, S.; Wang, F.; Li, Z.; and Sun, J. 2021.
\newblock Yolox: Exceeding yolo series in 2021.
\newblock \emph{arXiv preprint arXiv:2107.08430}.

\bibitem[{Gehrig et~al.(2019)Gehrig, Loquercio, Derpanis, and Scaramuzza}]{gehrig2019end}
Gehrig, D.; Loquercio, A.; Derpanis, K.~G.; and Scaramuzza, D. 2019.
\newblock End-to-end learning of representations for asynchronous event-based data.
\newblock In \emph{Proceedings of the IEEE/CVF International Conference on Computer Vision}, 5633--5643.

\bibitem[{Gehrig and Scaramuzza(2023)}]{gehrig2023recurrent}
Gehrig, M.; and Scaramuzza, D. 2023.
\newblock Recurrent vision transformers for object detection with event cameras.
\newblock In \emph{Proceedings of the IEEE/CVF conference on computer vision and pattern recognition}, 13884--13893.

\bibitem[{Hasssan, Meng, and Seo(2023)}]{hasssan2023lt}
Hasssan, A.; Meng, J.; and Seo, J.-s. 2023.
\newblock LT-SNN: Self-Adaptive Spiking Neural Network for Event-based Classification and Object Detection.

\bibitem[{Jiang et~al.(2019)Jiang, Xia, Huang, Stechele, Chen, Bing, and Knoll}]{jiang2019mixed}
Jiang, Z.; Xia, P.; Huang, K.; Stechele, W.; Chen, G.; Bing, Z.; and Knoll, A. 2019.
\newblock Mixed frame-/event-driven fast pedestrian detection.
\newblock In \emph{2019 International Conference on Robotics and Automation (ICRA)}, 8332--8338. IEEE.

\bibitem[{Lee et~al.(2025)Lee, Li, Kim, Xiao, and Panda}]{lee2024spiking}
Lee, D.; Li, Y.; Kim, Y.; Xiao, S.; and Panda, P. 2025.
\newblock Spiking Transformer with Spatial-Temporal Attention.
\newblock In \emph{Proceedings of the Computer Vision and Pattern Recognition Conference (CVPR)}, 13948--13958.

\bibitem[{Lee et~al.(2024)Lee, Yin, Kim, Moitra, Li, and Panda}]{lee2024tt}
Lee, D.; Yin, R.; Kim, Y.; Moitra, A.; Li, Y.; and Panda, P. 2024.
\newblock TT-SNN: tensor train decomposition for efficient spiking neural network training.
\newblock In \emph{2024 Design, Automation \& Test in Europe Conference \& Exhibition (DATE)}, 1--6. IEEE.

\bibitem[{Li et~al.(2022{\natexlab{a}})Li, Andreeto, Ranzato, and Perona}]{li2022caltech}
Li, F.; Andreeto, M.; Ranzato, M.; and Perona, P. 2022{\natexlab{a}}.
\newblock Caltech 101 (1.0)[Data set]. CaltechDATA.

\bibitem[{Li et~al.(2017)Li, Liu, Ji, Li, and Shi}]{li2017cifar10}
Li, H.; Liu, H.; Ji, X.; Li, G.; and Shi, L. 2017.
\newblock Cifar10-dvs: an event-stream dataset for object classification.
\newblock \emph{Frontiers in neuroscience}, 11: 309.

\bibitem[{Li et~al.(2022{\natexlab{b}})Li, Li, Zhu, Xiang, Huang, and Tian}]{li2022asynchronous}
Li, J.; Li, J.; Zhu, L.; Xiang, X.; Huang, T.; and Tian, Y. 2022{\natexlab{b}}.
\newblock Asynchronous spatio-temporal memory network for continuous event-based object detection.
\newblock \emph{IEEE Transactions on Image Processing}, 31: 2975--2987.

\bibitem[{Li et~al.(2021)Li, Guo, Zhang, Deng, Hai, and Gu}]{li2021differentiable}
Li, Y.; Guo, Y.; Zhang, S.; Deng, S.; Hai, Y.; and Gu, S. 2021.
\newblock Differentiable spike: Rethinking gradient-descent for training spiking neural networks.
\newblock \emph{Advances in neural information processing systems}, 34: 23426--23439.

\bibitem[{Li et~al.(2022{\natexlab{c}})Li, Kim, Park, Geller, and Panda}]{li2022neuromorphic}
Li, Y.; Kim, Y.; Park, H.; Geller, T.; and Panda, P. 2022{\natexlab{c}}.
\newblock Neuromorphic data augmentation for training spiking neural networks.
\newblock In \emph{European Conference on Computer Vision}, 631--649. Springer.

\bibitem[{Lichtsteiner, Posch, and Delbruck(2008)}]{lichtsteiner2008128}
Lichtsteiner, P.; Posch, C.; and Delbruck, T. 2008.
\newblock A 128$\times $128 120 dB 15$\mu $ s latency asynchronous temporal contrast vision sensor.
\newblock \emph{IEEE journal of solid-state circuits}, 43(2): 566--576.

\bibitem[{Liu and Delbruck(2018)}]{liu2018adaptive}
Liu, M.; and Delbruck, T. 2018.
\newblock Adaptive time-slice block-matching optical flow algorithm for dynamic vision sensors.
\newblock BMVC.

\bibitem[{Liu et~al.(2016)Liu, Anguelov, Erhan, Szegedy, Reed, Fu, and Berg}]{liu2016ssd}
Liu, W.; Anguelov, D.; Erhan, D.; Szegedy, C.; Reed, S.; Fu, C.-Y.; and Berg, A.~C. 2016.
\newblock Ssd: Single shot multibox detector.
\newblock In \emph{Computer Vision--ECCV 2016: 14th European Conference, Amsterdam, The Netherlands, October 11--14, 2016, Proceedings, Part I 14}, 21--37. Springer.

\bibitem[{Maass(1997)}]{maass1997networks}
Maass, W. 1997.
\newblock Networks of spiking neurons: the third generation of neural network models.
\newblock \emph{Neural networks}, 10(9): 1659--1671.

\bibitem[{Meng et~al.(2022)Meng, Xiao, Yan, Wang, Lin, and Luo}]{meng2022training}
Meng, Q.; Xiao, M.; Yan, S.; Wang, Y.; Lin, Z.; and Luo, Z.-Q. 2022.
\newblock Training high-performance low-latency spiking neural networks by differentiation on spike representation.
\newblock In \emph{Proceedings of the IEEE/CVF conference on computer vision and pattern recognition}, 12444--12453.

\bibitem[{Miao et~al.(2019)Miao, Chen, Ning, Zi, Ren, Bing, and Knoll}]{miao2019neuromorphic}
Miao, S.; Chen, G.; Ning, X.; Zi, Y.; Ren, K.; Bing, Z.; and Knoll, A.~C. 2019.
\newblock Neuromorphic Benchmark Datasets for Pedestrian Detection, Action Recognition, and Fall Detection.
\newblock \emph{Frontiers in neurorobotics}, 13: 38.

\bibitem[{Orchard et~al.(2015)Orchard, Jayawant, Cohen, and Thakor}]{orchard2015converting}
Orchard, G.; Jayawant, A.; Cohen, G.~K.; and Thakor, N. 2015.
\newblock Converting static image datasets to spiking neuromorphic datasets using saccades.
\newblock \emph{Frontiers in neuroscience}, 9: 437.

\bibitem[{Park and Kim(2022)}]{park2022vision}
Park, N.; and Kim, S. 2022.
\newblock How do vision transformers work?
\newblock \emph{arXiv preprint arXiv:2202.06709}.

\bibitem[{Perot et~al.(2020{\natexlab{a}})Perot, de~Tournemire, Nitti, Masci, and Sironi}]{perot2020learningdetectobjects1}
Perot, E.; de~Tournemire, P.; Nitti, D.; Masci, J.; and Sironi, A. 2020{\natexlab{a}}.
\newblock Learning to Detect Objects with a 1 Megapixel Event Camera.
\newblock \emph{arXiv preprint arXiv:2009.13436}.

\bibitem[{Perot et~al.(2020{\natexlab{b}})Perot, De~Tournemire, Nitti, Masci, and Sironi}]{perot2020learning}
Perot, E.; De~Tournemire, P.; Nitti, D.; Masci, J.; and Sironi, A. 2020{\natexlab{b}}.
\newblock Learning to detect objects with a 1 megapixel event camera.
\newblock \emph{Advances in Neural Information Processing Systems}, 33: 16639--16652.

\bibitem[{Posch, Matolin, and Wohlgenannt(2010)}]{posch2010qvga}
Posch, C.; Matolin, D.; and Wohlgenannt, R. 2010.
\newblock A QVGA 143 dB dynamic range frame-free PWM image sensor with lossless pixel-level video compression and time-domain CDS.
\newblock \emph{IEEE Journal of Solid-State Circuits}, 46(1): 259--275.

\bibitem[{Roy, Jaiswal, and Panda(2019)}]{roy2019towards}
Roy, K.; Jaiswal, A.; and Panda, P. 2019.
\newblock Towards spike-based machine intelligence with neuromorphic computing.
\newblock \emph{Nature}, 575(7784): 607--617.

\bibitem[{Shi, Hao, and Yu(2024)}]{shi2024spikingresformer}
Shi, X.; Hao, Z.; and Yu, Z. 2024.
\newblock SpikingResformer: bridging ResNet and vision transformer in spiking neural networks.
\newblock In \emph{Proceedings of the IEEE/CVF Conference on Computer Vision and Pattern Recognition}, 5610--5619.

\bibitem[{Sironi et~al.(2018)Sironi, Brambilla, Bourdis, Lagorce, and Benosman}]{sironi2018hats}
Sironi, A.; Brambilla, M.; Bourdis, N.; Lagorce, X.; and Benosman, R. 2018.
\newblock HATS: Histograms of averaged time surfaces for robust event-based object classification.
\newblock In \emph{Proceedings of the IEEE conference on computer vision and pattern recognition}, 1731--1740.

\bibitem[{Su et~al.(2023)Su, Chou, Hu, Li, Mei, Zhang, and Li}]{su2023deep}
Su, Q.; Chou, Y.; Hu, Y.; Li, J.; Mei, S.; Zhang, Z.; and Li, G. 2023.
\newblock Deep directly-trained spiking neural networks for object detection.
\newblock In \emph{Proceedings of the IEEE/CVF International Conference on Computer Vision}, 6555--6565.

\bibitem[{Vaswani et~al.(2017)Vaswani, Shazeer, Parmar, Uszkoreit, Jones, Gomez, Kaiser, and Polosukhin}]{vaswani2017attention}
Vaswani, A.; Shazeer, N.; Parmar, N.; Uszkoreit, J.; Jones, L.; Gomez, A.~N.; Kaiser, {\L}.; and Polosukhin, I. 2017.
\newblock Attention is all you need.
\newblock \emph{Advances in neural information processing systems}, 30.

\bibitem[{Wang et~al.(2022{\natexlab{a}})Wang, Zheng, Chen, and Wang}]{wang2022anti}
Wang, P.; Zheng, W.; Chen, T.; and Wang, Z. 2022{\natexlab{a}}.
\newblock Anti-oversmoothing in deep vision transformers via the fourier domain analysis: From theory to practice.
\newblock \emph{arXiv preprint arXiv:2203.05962}.

\bibitem[{Wang et~al.(2022{\natexlab{b}})Wang, Luo, Wang, Ding, Wang, and Li}]{wang2022vtc}
Wang, Z.; Luo, H.; Wang, P.; Ding, F.; Wang, F.; and Li, H. 2022{\natexlab{b}}.
\newblock Vtc-lfc: Vision transformer compression with low-frequency components.
\newblock \emph{Advances in Neural Information Processing Systems}, 35: 13974--13988.

\bibitem[{Wang et~al.(2024)Wang, Wang, Li, Qin, Jiang, Ma, and Tang}]{wang2024eas}
Wang, Z.; Wang, Z.; Li, H.; Qin, L.; Jiang, R.; Ma, D.; and Tang, H. 2024.
\newblock Eas-snn: End-to-end adaptive sampling and representation for event-based detection with recurrent spiking neural networks.
\newblock In \emph{European Conference on Computer Vision}, 310--328. Springer.

\bibitem[{Yao et~al.(2024{\natexlab{a}})Yao, Hu, Hu, Xu, Zhou, Tian, Xu, and Li}]{yao2024spike2}
Yao, M.; Hu, J.; Hu, T.; Xu, Y.; Zhou, Z.; Tian, Y.; Xu, B.; and Li, G. 2024{\natexlab{a}}.
\newblock Spike-driven transformer v2: Meta spiking neural network architecture inspiring the design of next-generation neuromorphic chips.
\newblock \emph{arXiv preprint arXiv:2404.03663}.

\bibitem[{Yao et~al.(2023)Yao, Hu, Zhou, Yuan, Tian, Xu, and Li}]{yao2023spike}
Yao, M.; Hu, J.; Zhou, Z.; Yuan, L.; Tian, Y.; Xu, B.; and Li, G. 2023.
\newblock Spike-driven transformer.
\newblock \emph{Advances in neural information processing systems}, 36: 64043--64058.

\bibitem[{Yao et~al.(2024{\natexlab{b}})Yao, Hu, Zhou, Yuan, Tian, Xu, and Li}]{yao2024spike}
Yao, M.; Hu, J.; Zhou, Z.; Yuan, L.; Tian, Y.; Xu, B.; and Li, G. 2024{\natexlab{b}}.
\newblock Spike-driven transformer.
\newblock \emph{Advances in Neural Information Processing Systems}, 36.

\bibitem[{Yu et~al.(2024)Yu, Chen, Wang, Zhan, Shao, Liu, and Xu}]{yu2024spikingvit}
Yu, L.; Chen, H.; Wang, Z.; Zhan, S.; Shao, J.; Liu, Q.; and Xu, S. 2024.
\newblock Spikingvit: a multi-scale spiking vision transformer model for event-based object detection.
\newblock \emph{IEEE Transactions on Cognitive and Developmental Systems}.

\bibitem[{Yun et~al.(2019)Yun, Han, Oh, Chun, Choe, and Yoo}]{yun2019cutmix}
Yun, S.; Han, D.; Oh, S.~J.; Chun, S.; Choe, J.; and Yoo, Y. 2019.
\newblock Cutmix: Regularization strategy to train strong classifiers with localizable features.
\newblock In \emph{Proceedings of the IEEE/CVF international conference on computer vision}, 6023--6032.

\bibitem[{Zhang et~al.(2022)Zhang, Dong, Zhang, Ding, Heide, Yin, and Yang}]{zhang2022spiking}
Zhang, J.; Dong, B.; Zhang, H.; Ding, J.; Heide, F.; Yin, B.; and Yang, X. 2022.
\newblock Spiking transformers for event-based single object tracking.
\newblock In \emph{Proceedings of the IEEE/CVF conference on Computer Vision and Pattern Recognition}, 8801--8810.

\bibitem[{Zheng et~al.(2021)Zheng, Wu, Deng, Hu, and Li}]{zheng2021going}
Zheng, H.; Wu, Y.; Deng, L.; Hu, Y.; and Li, G. 2021.
\newblock Going deeper with directly-trained larger spiking neural networks.
\newblock In \emph{Proceedings of the AAAI conference on artificial intelligence}, volume~35, 11062--11070.

\bibitem[{Zhou et~al.(2024)Zhou, Zhang, Zhou, Yu, Huang, Fan, Yuan, Ma, Zhou, and Tian}]{zhou2024qkformer}
Zhou, C.; Zhang, H.; Zhou, Z.; Yu, L.; Huang, L.; Fan, X.; Yuan, L.; Ma, Z.; Zhou, H.; and Tian, Y. 2024.
\newblock Qkformer: Hierarchical spiking transformer using qk attention.
\newblock \emph{arXiv preprint arXiv:2403.16552}.

\bibitem[{Zhou et~al.(2022)Zhou, Zhu, He, Wang, Yan, Tian, and Yuan}]{zhou2022spikformer}
Zhou, Z.; Zhu, Y.; He, C.; Wang, Y.; Yan, S.; Tian, Y.; and Yuan, L. 2022.
\newblock Spikformer: When spiking neural network meets transformer.
\newblock \emph{arXiv preprint arXiv:2209.15425}.

\end{thebibliography}

\clearpage
\appendix
\section{A. Experimental Details}
\label{ex_details}
This section details the experimental setup for classification tasks (CIFAR10-DVS, N-Caltech101) and object detection tasks (PAF, Gen1).

\begin{table}[h]
\footnotesize
  \caption{Experimental configuration for classification tasks.}
  \label{appendix_exp_classification}
  \centering
  \begin{tabular}{ccc}
    \toprule
    & CIFAR10-DVS & N-Caltech101  \\
    \midrule
    Timestep & 16 &  16  \\
    Batch size & 16 &  16\\ 
    Learning rate & 0.01 & 0.01\\ 
    Training epochs & 210  & 210\\ 
    Optimizer & AdamW & AdamW \\
    Hardware (GPU) & A5000 & A5000\\ 
    \bottomrule
  \end{tabular}
\end{table}

\noindent \textbf{Classification. }We adopt the Spike-driven Transformer~\cite{yao2023spike} configuration as our baseline and modify it for our SpikePool architecture as shown in Table~\ref{appendix_exp_classification}. For data augmentation, CIFAR10-DVS frames are resized to 64$\times$64 pixels using the SpikingJelly~\cite{fang2023spikingjelly} dataloader. For N-Caltech101, we follow the NDA~\cite{li2022neuromorphic} augmentation protocol, which includes horizontal flip, rolling, rotation, and CutMix~\cite{yun2019cutmix} techniques.

\noindent \textbf{Object Detection.} We train both SpikePool and SpikingViT~\cite{yu2024spikingvit} on the Gen1 and 1Mpx datasets. For Gen1, we use a batch size of 24, sequence length of 5, and initial learning rate of 2e–4, training for 300,000 iterations over 4-5 days on a single A100 GPU. For 1Mpx, we employ a batch size of 10, sequence length of 5, and initial learning rate of 3e–4, training for 500,000 iterations over 11-12 days on the same hardware. We utilize mixed precision training with the Adam optimizer for both datasets.

For the PAF experiments, we split the full PAF dataset into an 80\% training set and a 20\% validation set. Image frames are 260$\times$346 pixels and each 20 ms event window is discretized into 20 temporal bins for voxelization. Training is performed for 10,000 iterations on a single A5000 GPU with a batch size of 16, initial learning rate of 3e-4, and also utilizes mixed precision training with the Adam optimizer. Under these settings the training finishes in approximately 2 days.

\begin{figure}[!ht]
  \centering
  \includegraphics[width=0.80\linewidth]{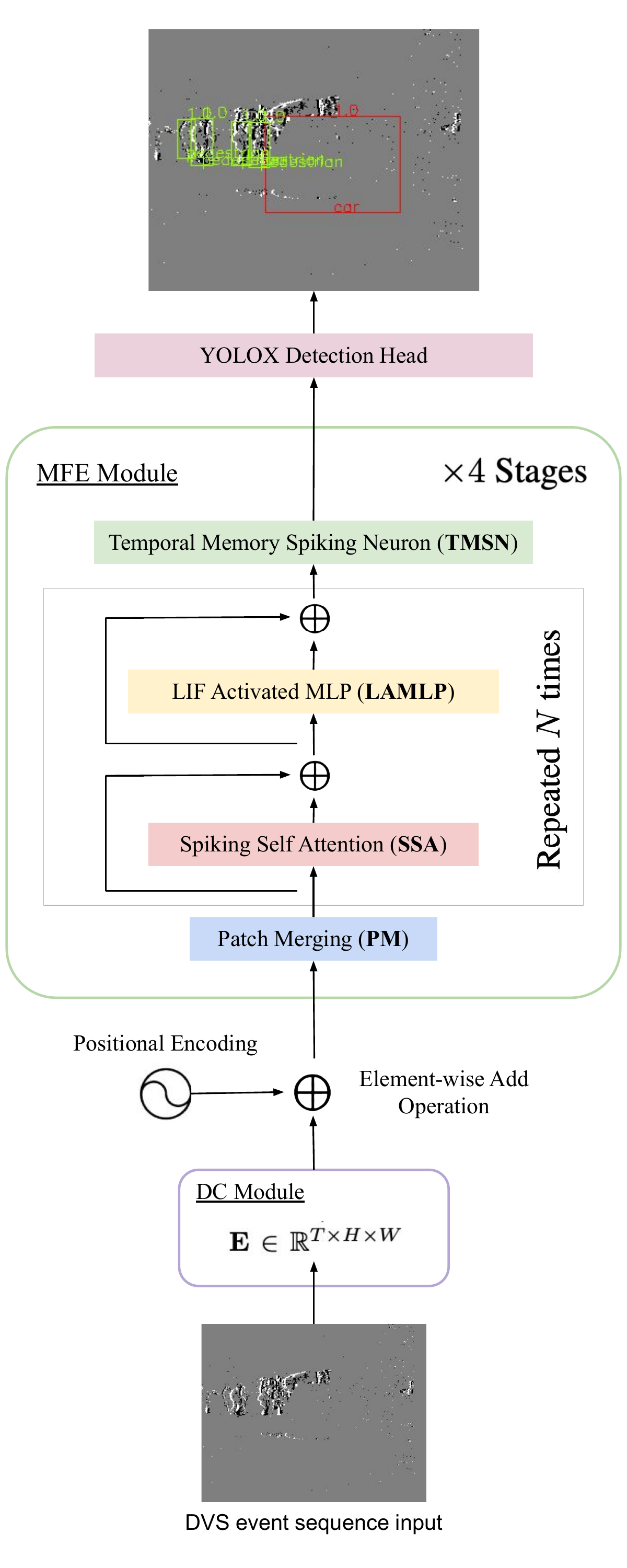}
  \vspace{-6pt}
  \caption{Overview of the SpikingViT architecture for event-based object detection. For our SpikePool experiment, we replaced the Spiking Self-Attention (SSA) block shown above with max-pooling.}
  \label{fig:spikevit_viz}
  \vspace{-8pt}
\end{figure}

\section{B. SpikingViT Architecture}
\label{s_vit}

The SpikingViT~\cite{yu2024spikingvit} integrates SNNs with transformer architectures to achieve energy-efficient yet high-performance visual processing. SpikingViT introduces several core components, including its Dynamic Convolution (DC) module, Multi-scale Feature Extraction (MFE) mechanism, and specialized blocks such as LIF-Activated Multilayer Perceptron (LAMLP) and Temporal Memory Spiking Neuron (TMSN), culminating in its integration with a YOLOX~\cite{ge2021yolox} detection head for object detection.

\subsection{DC Module}
The DC module quantizes the DVS event sequence for each time slot into tensors containing comprehensive spatiotemporal information (this results in a 4D tensor $\mathbf{E} \in \mathbb{R}^{2 \times T \times H \times W}$ where $T$, $H$, and $W$ denote the aggregation time, and preprocessed height and width, respectively). Notably, ~\cite{yu2024spikingvit} combines the first dimension, which represents polarity, to obtain a new tensor $\mathbf{E} \in \mathbb{R}^{T\times H\times W}$, reducing escalations in memory usage and magnified fluctuations in gradients due to preservation of positive and negative polarity channels.

\subsection{MFE Module}
Upon completion of data preprocessing, the MFE module facilitates extraction and learning of object features.

\noindent \textbf{Patch Merging Block.} Initially, the feature map obtained from DC module is fed into the PM block, which serves two main functions in each learning stage: spatial down‐sampling (merging adjacent patches) and temporal expansion (increasing the number of timesteps). In each of its learning stages, the PM block concatenates features from each group of 2$\times$2 neighboring patches, resulting in a down-sampling of the tensor resolution by a factor of 2 in both height and width dimensions. Simultaneously, a small linear layer on the concatenated features expands the timestep dimension from $T$ to $2T$. In other words, the PM block reshapes: $$[T, H, W, C] \xrightarrow{\text{merge + project}}\left[2T, \frac{H}{2}, \frac{W}{2}, C^{'}\right],$$ which ensures that each subsequent SSA block processes a longer spike‐train sequence, enhancing temporal integration, while allowing the model to extract multi‐scale spatio‐temporal features.  As a result, the timestep $T$ in each stage is variable. When the length of the timestep doubles, the corresponding resolution is halved. This mechanism aims to extract feature information at different scales more effectively. Output from the PM block is then fed into SSA, which is shown in Section~\ref{ssa}. Notably, ~\cite{yu2024spikingvit} employs nonfixed scaling factors to regulate the magnitudes of matrix multiplication results, as opposed to the fixed scaling factors employed in ~\cite{zhou2022spikformer}.

\subsection{LAMLP and TMSN}

After the SSA block, the LAMLP block employs LIF spike neurons to construct a multilayer perceptron block, which captures hidden correlation information from both temporal and spatial dimensions while incorporating additional learnable parameters. The TMSN block is introduced at the end of each learning stage, which exploits residual voltage memory created by LIF neurons to enhance model performance. Finally, the output feature maps are fed into the YOLOX detection head for object detection. An overview of the entire SpikingViT architecture is shown in Figure ~\ref{fig:spikevit_viz}.

\section{C. Generalization Implementation}
For architectural generalization, we conduct plug-in experiments with pooling attention on existing spiking transformer architectures: Spikformer~\cite{zhou2022spikformer} and QKFormer~\cite{zhou2024qkformer}, as shown in Table~\ref{tab:plug}. In this section, we detail the architectural modifications required to integrate our pooling attention into these frameworks.

\noindent\textbf{Spikformer:} Spikformer is the first proposed spiking transformer architecture featuring the SSA module. We directly replace the SSA component with our max pooling operation, eliminating QKV computations entirely. This modification yields consistent performance improvements of 0.3\% and 0.87\% on CIFAR10-DVS and N-Caltech101 datasets, respectively.

\noindent\textbf{QKFormer:} QKFormer employs two distinct self-attention mechanisms, Q-K Token Attention (QKTA) and SSA. Our experiments reveal that the optimal attention replacement varies by dataset. For CIFAR10-DVS, we achieve the best results by replacing QKTA, while for N-Caltech101, replacing SSA yields superior performance. This suggests that pooling attention can be selectively applied based on the specific architectural design and dataset characteristics, while maintaining its general applicability across different spiking transformer frameworks.

All training experiments follow the default hyperparameter settings of their respective original implementations to ensure fair comparison.

\section{D. SpikePool on RGB dataset}

To evaluate the generalizability of our SpikePool architecture beyond event-based data, we conduct experiments on standard RGB image classification datasets. 
\begin{table}[h]
\footnotesize
  \centering
  \caption{Accuracy comparison on RGB dataset, CIFAR10 and CIFAR100 dataset.}
  \label{tab:rgb}
  \begin{tabular}{lcc}
    \toprule
    Model & CIFAR10 (\%) & CIFAR100 (\%) \\
    \midrule
    Spike-driven Transformer & 95.6 & 78.4 \\
    SpikePool & 95.24 & 78.05\\
    \bottomrule
  \end{tabular}
\end{table}

Table~\ref{tab:rgb} presents the accuracy comparison between our SpikePool and the Spike-driven Transformer~\cite{yao2023spike} on CIFAR-10 and CIFAR-100 datasets. The results show comparable performance, with slight differences within experimental variance (CIFAR-10: 0.36\% $\downarrow$, CIFAR-100: 0.35\% $\downarrow$). This indicates that while SpikePool does not provide the significant advantages observed on event-based datasets, it maintains competitive performance on RGB data. 

The modest performance difference on RGB datasets, compared to substantial improvements on event-based data, highlights the domain-specific nature of our approach. RGB images lack the natural spatial-temporal sparsity and dynamics that make max pooling particularly effective. Instead, these datasets benefit more from the complex spatial relationship modeling provided by QKV self-attention mechanisms. This domain specificity confirms that SpikePool is purpose-built for neuromorphic data: our frequency balancing through max pooling optimally exploits sparse temporal dynamics inherent in event streams, not dense spatial patterns in RGB images.

\onecolumn
\section{E. Additional Visualization}
\vspace{-5pt}
\subsection{Failure Case on Gen1 Dataset}
\label{add_vis}
\vspace{-5pt}

\begin{figure}[!htbp]
  \centering
  \begin{tabular}{@{}c@{\hspace{4pt}}c@{\hspace{4pt}}c@{\hspace{4pt}}c@{\hspace{4pt}}c@{}}
    \begin{minipage}[b]{.03\linewidth}
      \rotatebox{90}{\footnotesize SpikingViT}\\[50pt]
      \rotatebox{90}{\footnotesize SpikePool}\\[50pt]
      \rotatebox{90}{\footnotesize Ground Truth}\\[20pt]
    \end{minipage}
    \begin{subfigure}[b]{0.205\linewidth}
      \includegraphics[width=\linewidth]{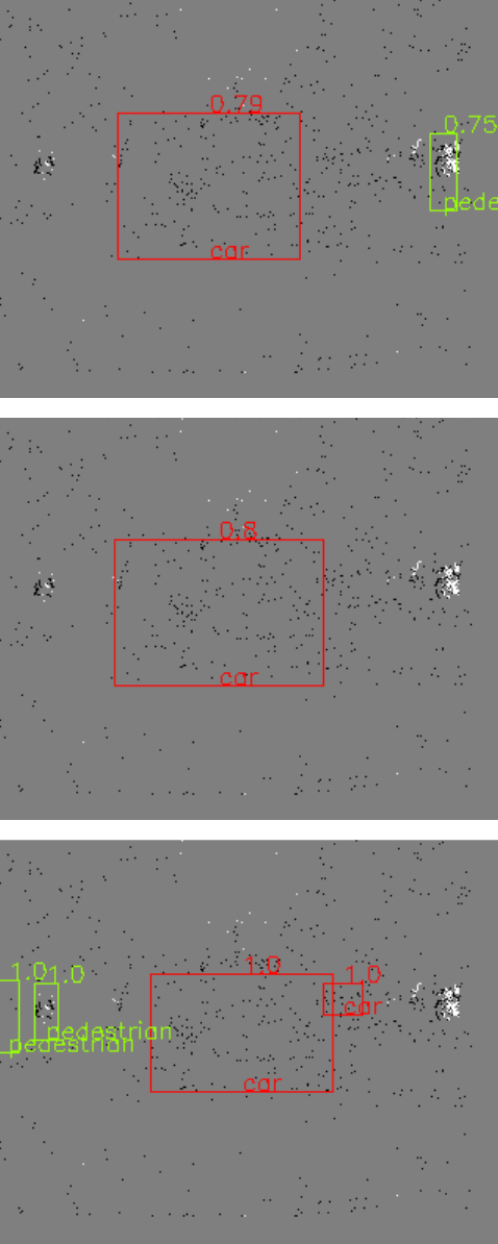}
      \caption{Example 1}
      \label{fig:spikepool_fail_1}
    \end{subfigure} &
    \begin{subfigure}[b]{0.205\linewidth}
      \includegraphics[width=\linewidth]{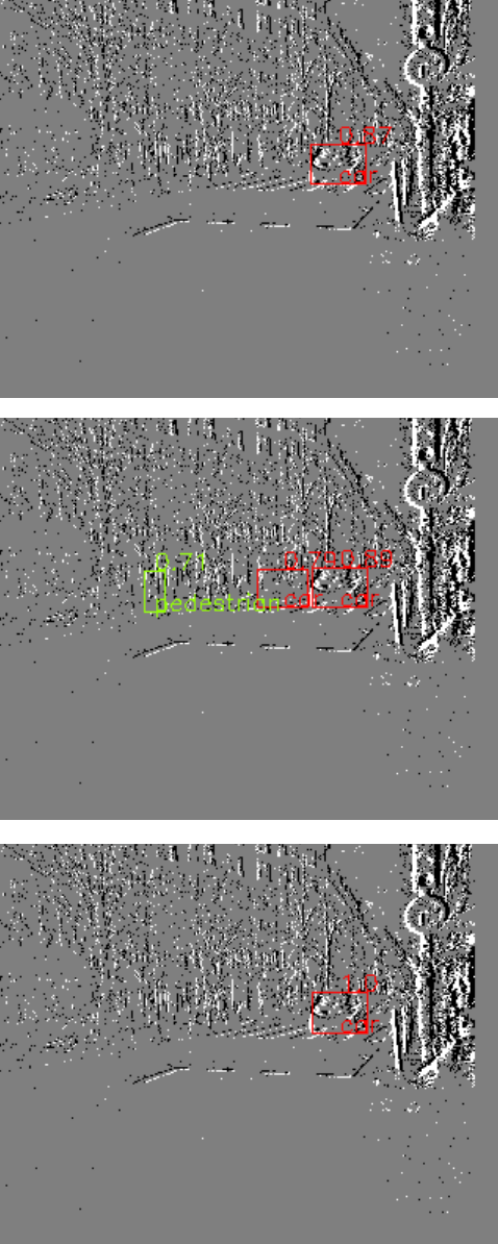}
      \caption{Example 2}
      \label{fig:spikepool_fail_2}
    \end{subfigure} &
    \begin{subfigure}[b]{0.205\linewidth}
      \includegraphics[width=\linewidth]{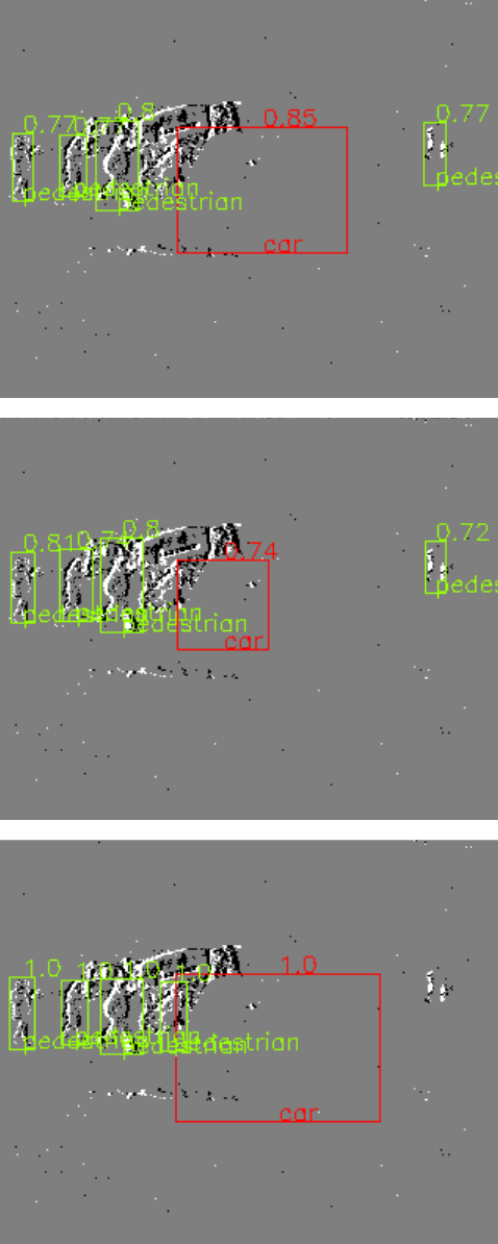}
      \caption{Example 3}
      \label{fig:spikepool_fail_3}
    \end{subfigure}
  \end{tabular}

  \vspace{-8pt}
  \caption{SpikePool, SpikingViT, and Ground Truth visualizations on failure cases in the Gen1 dataset. In Example 1, SpikePool is unable to identify any pedestrians. In Example 2, SpikePool identifies an extra car and pedestrian. In Example 3, SpikePool's bounding box for the car is too thin.}
  \label{fig:spikepool_fail_viz}
\end{figure}

\vspace{-15pt}

\subsection{Visualizations on PAF Dataset}

\begin{figure}[!htbp]
  \centering
  \begin{tabular}{@{}c@{\hspace{4pt}}c@{\hspace{4pt}}c@{\hspace{4pt}}c@{\hspace{4pt}}c@{}}
    \begin{minipage}[b]{.03\linewidth}
      \rotatebox{90}{\footnotesize SpikingViT}\\[45pt]
      \rotatebox{90}{\footnotesize SpikePool}\\[45pt]
      \rotatebox{90}{\footnotesize Ground Truth}\\[17pt]
    \end{minipage}
    \begin{subfigure}[b]{0.205\linewidth}
      \includegraphics[width=\linewidth]{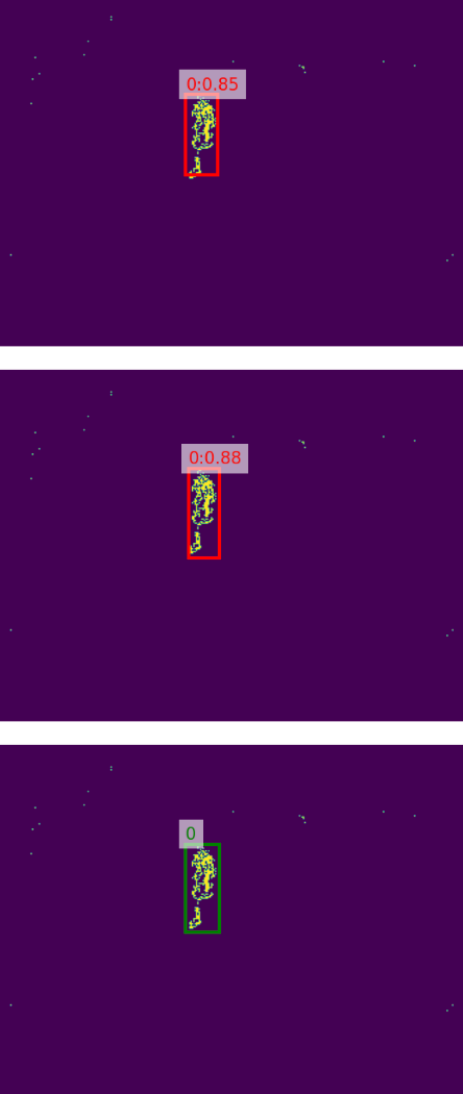}
      \caption{Example 1}
      \label{fig:spikepool_paf_1}
    \end{subfigure} &
    \begin{subfigure}[b]{0.205\linewidth}
      \includegraphics[width=\linewidth]{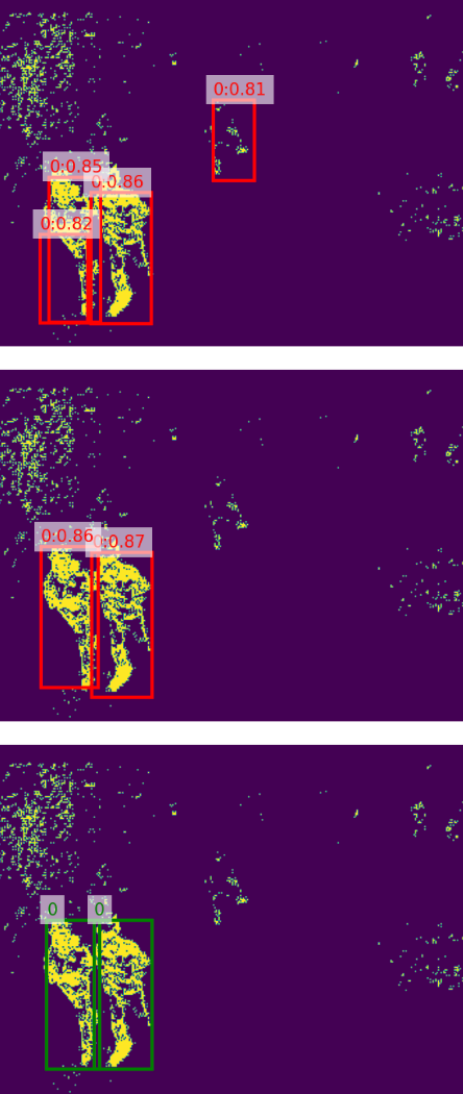}
      \caption{Example 2}
      \label{fig:spikepool_paf_2}
    \end{subfigure} &
    \begin{subfigure}[b]{0.205\linewidth}
      \includegraphics[width=\linewidth]{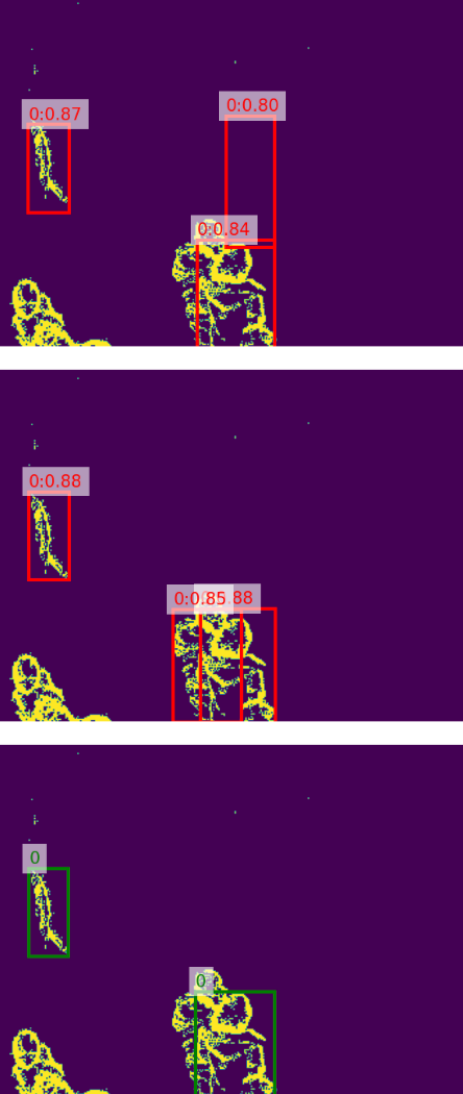}
      \caption{Example 3}
      \label{fig:spikepool_paf_3}
    \end{subfigure}
  \end{tabular}

  \vspace{-6pt}
  \caption{SpikePool, SpikingViT, and Ground Truth visualizations on the PAFBenchmark dataset.}
  \label{fig:spikepool_paf_viz}
  \vspace{-8pt}
\end{figure}

\end{document}